\documentclass[fleqn,10pt]{wlscirep}
\usepackage[utf8]{inputenc}
\usepackage[T1]{fontenc}
\usepackage{physics}
\usepackage{algorithm2e}
\usepackage{cancel}

\usepackage{graphicx}
\usepackage{booktabs}

\usepackage{amsmath,amsfonts,amssymb}
\usepackage{verbatim}
\DeclareMathOperator*{\argmin}{argmin}

\usepackage{bm} 

\usepackage{float}
\usepackage{color,soul}

\def\R{{\mathbb R}} \def\C{{\mathbb C}} 
\def\Z{{\mathbb Z}} \def\one{\mbox{1 \kern-.59em {\rm l}}}

\def\cM{{\cal M}}

\def\Mat{\rm Mat}

\usepackage[font=footnotesize, labelfont=bf, justification=justified]{caption}
\usepackage{subcaption}
\usepackage{lineno}

\usepackage{hyperref}
\hypersetup{hidelinks}

\title{Quantum Geometry of Data}

\author[1,2,*]{Alexander G. Abanov}
\author[1,3]{Luca Candelori}
\author[1,4]{Harold C. Steinacker}
\author[1,5]{Martin T. Wells}

\author[1,6,+]{Jerome R. Busemeyer}
\author[5,+]{Cameron J. Hogan}
\author[1,+]{Vahagn Kirakosyan}
\author[7,+]{Nicola Marzari}
\author[1,+]{Sunil Pinnamaneni}
\author[1,8,+]{Dario Villani}
\author[1,9,10,+]{Mengjia Xu}

\affil[1]{Qognitive, Inc., Miami Beach, FL 33139, USA}
\affil[2]{Stony Brook University, Department of Physics and Astronomy, Stony Brook, NY 11790, USA}
\affil[3]{Wayne State University, Department of Mathematics, Detroit, MI 48202, USA}
\affil[4]{University of Vienna, Faculty of Physics, Boltzmanngasse 5, A-1090 Vienna, Austria}
\affil[5]{Cornell University, Department of Statistics and Data Science, Ithaca, NY 14853, USA}
\affil[6]{Department of Psychological and Brain Sciences, Indiana University.}

\affil[7]{Theory and Simulations of Materials (THEOS),
and National Centre for Computational Design and Discovery of Novel Materials (MARVEL),
\'Ecole Polytechnique F\'ed\'erale de Lausanne, 1015 Lausanne, Switzerland}

\affil[8]{King’s College London, Department of Mathematics, London, UK}
\affil[9]{New Jersey Institute of Technology, Department of Data Science, Newark, NJ 07102, USA}
\affil[10]{Massachusetts Institute of Technology, Center for Brains, Minds and Machines, Cambridge, MA 02139, USA}

\author[1]{Kharen Musaelian}

\affil[*]{abanov@qognitive.io, alexandre.abanov@stonybrook.edu}
\affil[+]{these authors contributed equally to this work and are listed alphabetically}

\keywords{Quantum Geometry, Matrix Laplacian, Quantum Cognition Machine Learning (QCML), Dimension Reduction, Spectral Methods, Fuzzy Manifolds, Data Science}

\begin{abstract}
We demonstrate how Quantum Cognition Machine Learning (QCML) encodes data as quantum geometry. 
In QCML, features of the data are represented by learned Hermitian matrices, and data points are mapped to states in Hilbert space. 
The quantum geometry description endows the dataset with rich geometric and topological structure---including intrinsic dimension, quantum metric, and Berry curvature---derived directly from the data. 
QCML captures global properties of data through a compact operator representation. At fixed Hilbert-space dimension, the number of operator parameters grows only linearly with the number of features. 
The present work provides the first full exposition of QCML as a framework for \emph{matrix geometry}, 
establishing its mathematical foundation and demonstrating how operator-level tools from quantum geometry can be directly applied to data. 
We introduce the matrix Laplacian and its eigenmaps as a geometry-preserving alternative to graph-based embeddings, 
and we show how topological invariants such as Chern numbers can be extracted from data for the first time. 
An important practical outcome is the efficiency of the representation:  the relevant geometric structure of high-dimensional datasets can be captured with Hilbert spaces of size as small as $N \sim 10$. 
{ We illustrate these results on a number of synthetic and real-world examples, including the Wisconsin Breast Cancer dataset, 
where QCML uncovers geometric separability between benign and malignant samples and yields an interpretable decomposition of feature relevance.}  

The emphasis of this work is on a new geometric representation of data rather than on algorithmic benchmarking, establishing a foundation for extracting global geometric and topological structure from high-dimensional datasets.

\end{abstract}

\begin{document}

\flushbottom
\maketitle

\tableofcontents

\thispagestyle{empty}

\section{Introduction}

In data analysis, one often has to work with data involving a large number of features. As an example, consider patient records containing vitals, RNA sequencing, images, etc. In this case, good sampling of the distribution requires a number of data points that grows exponentially with the number of features---a problem often referred to as the curse of dimensionality \cite{Bellman1957}. However, in many cases, only a few features (or their combinations) are relevant to the task at hand, which results in a concentration of measure---the data points concentrate near a manifold with relatively low intrinsic dimension.

Discovering the underlying data manifold and learning its properties is the central goal of manifold learning. Substantial progress has been made in the development of manifold learning methods \cite{tenenbaum2000global,roweis2000nonlinear,belkin2003laplacian,DonohoGrimes2003}. However, these methods typically rely on local neighborhood structures and are not effective in capturing global properties of the dataset. Although there have been advances in this direction, for example in topological data analysis \cite{wasserman2018topological}, the problem of understanding high-dimensional datasets remains central.

Recently, a new approach to data analysis, quantum cognition machine learning (QCML), has been introduced \cite{musaelian2024qcml,candelori2025robust}. In QCML, data points are represented as quantum states (i.e., vectors in a finite \( N \)-dimensional Hilbert space), and features or variables are represented by observables (i.e., Hermitian matrices acting on the states). Both quantum states and observables are jointly learned through a machine learning training process, resulting in a new representation of the data. Applications of QCML have demonstrated that important features of various datasets—ranging from medical to financial and synthetic \cite{candelori2025robust,di2025quantum,samson2024quantum,rosaler2025supervised}—can be captured with modest Hilbert space sizes, such as \( 8 \)–\( 32 \). This shows how the curse of dimensionality can be overcome in cases where important insights can be achieved with relatively small values of \( N \). Heuristically, one can think of \( N \) as a magnification factor: larger values allow for resolving finer details in the data, while smaller \( N \) result in a more coarse-grained, smeared representation.

An important feature of QCML, which we emphasize in this work, is its \emph{efficiency}: 
 relevant geometric structure in datasets with dozens of features can be captured using 
Hilbert spaces of dimension as small as \( N \sim 10 \). 
This remarkable compression indicates that QCML captures 
the essential nonlinear structure of data manifolds. 
Efficiency is crucial for scaling to scientific and biomedical applications involving large and complex datasets.

In this work, we explore how data encoding in QCML is related to quantum geometry, as understood in mathematics and theoretical physics. 
Specifically, the set of learned matrices can be interpreted as defining a quantum or matrix geometry in the sense described in Refs.~\citeonline{steinacker2024quantum,steinacker2021quantum,ishiki2015matrix}; see Section~\ref{sec:QG} for a brief review. 
While earlier QCML works focused on specific applications, the present paper provides the first full exposition of QCML as a framework for \emph{matrix geometry}, establishing the mathematical foundation that connects learned operator algebras with fuzzy or noncommutative spaces. 
It is therefore natural to scrutinize the data-induced quantum geometry obtained by QCML for model datasets and compare it with insights from quantum geometry.

Beyond establishing this conceptual bridge, 
we also demonstrate the \emph{practical consequences} of the novel tools and methods provided by this framework: 
computing intrinsic dimension, extracting geometric and topological invariants, 
and constructing embeddings that preserve global data structure. 
These diagnostics provide interpretable insights, 
for example into { the geometric organization of diagnostic classes and the relative contribution of individual features in biomedical datasets.}

We obtain quantum geometry from QCML training and then apply a range of techniques developed in the quantum geometry literature for data analysis. We demonstrate that, for geometric synthetic datasets, QCML effectively learns the quantum geometry of the corresponding geometric objects. We extract various properties of these objects, such as connectivity, the quantum geometric tensor, Chern numbers, and spectra of matrix Laplacians. We also consider an example of a real-life dataset---the Diagnostic Wisconsin Breast Cancer Database \cite{misc_breast_cancer_wisconsin_(diagnostic)_17}.  In this case, where no \textit{a priori} geometric knowledge is available, the quantum geometry analysis provides valuable insight into the structure of the dataset.

The key contributions of this work can be summarized as follows:
\begin{enumerate}
  \item We present the first comprehensive formulation of QCML as a general framework, 
  showing that it naturally defines a \emph{matrix (fuzzy) geometry} in which 
  data points correspond to quasi-coherent states and features to Hermitian operators. 

  \item From the learned QCML representation, we compute fundamental geometric quantities—the 
  \emph{quantum geometric tensor}, \emph{Berry curvature}, and \emph{Chern numbers}—and introduce 
  the \emph{matrix Laplacian} with its eigenmaps as a geometry-preserving analogue of graph Laplacians. 

  \item We demonstrate that  relevant geometric structure in high-dimensional datasets can be represented within Hilbert spaces of dimension as small as \(N \sim 10\), 
  revealing a remarkable compactness and scalability of the approach. 

  \item { Through applications to both synthetic datasets (disconnected clusters, conformal maps) and the 
  Wisconsin Breast Cancer Database, we show how QCML provides global embeddings, curvature- and 
  dimension-based diagnostics.} 

  \item By unifying concepts from quantum geometry and data science, QCML establishes a new operator-based 
  paradigm for data analysis with broad implications across physics, mathematics, and machine learning.
\end{enumerate}

For clarity, we emphasize that the primary goal of this work is \emph{not} to introduce a new machine-learning algorithm to be benchmarked against existing classical methods, nor to claim immediate performance advantages on standard tasks such as classification or clustering. 
Rather, our objective is to develop a \emph{new representation of data}: an operator-based encoding in which high-dimensional datasets are mapped to finite-dimensional quantum (matrix) geometries.
This representation enables the extraction of geometric and topological structures--such as intrinsic dimension, curvature, and Chern numbers--that are not naturally accessible within conventional graph-based or linear frameworks.

In this context, \emph{efficiency} refers to representational compactness rather than raw computational speed.
 In our experience across a range of datasets, relevant structure has often been captured, even at very large feature dimension, using Hilbert spaces of dimension as small as $N \sim 10$.
{ For fixed Hilbert-space dimension \(N\) and sample size \(T\), both the number of learned operator parameters and the dominant training cost per epoch scale linearly with the ambient data dimension \(D\), avoiding exponential dependence on \(D\).}
Here, avoidance of the curse of dimensionality refers to the size of
the operator representation for datasets whose relevant structure can be
captured at small \(N\).
This small-$N$ regime is a deliberate design choice, reflecting the aim of capturing \emph{global nonlinear structure} with a minimal number of degrees of freedom.
Systematic benchmarking against classical dimensionality-reduction techniques is an important direction for future work, but lies beyond the scope of the present foundational study.

This article is organized as follows. We begin with a brief overview of the connection between data and quantum geometry in the context of manifold learning. Section~\ref{sec:QCML} outlines the quantum cognition machine learning (QCML) framework.
To fully develop this framework, we review essential concepts from quantum geometry in Section~\ref{sec:QG}, particularly the use of displacement Hamiltonians and quasi-coherent states. 
In Section~\ref{sec:topological}, we show how more refined geometric information—such as the quantum geometric tensor and integer-valued topological invariants—can be extracted from quantum geometry. 
One of the most important tools for quantum geometry analysis, the matrix Laplacian, is introduced in Section~\ref{sec:matrix-Laplacian}. In particular, it can be used to find matrix Laplacian eigenmaps that are instrumental in constructing matrix representations which optimally preserve geometric information (Section~\ref{sec:eigenmaps}). We also discuss how quantum geometry overcomes the curse of dimensionality by efficiently representing high-dimensional structures with relatively few parameters (Section~\ref{sec:curse}).

In Section~\ref{sec:examples}, we apply the introduced techniques to a range of illustrative examples, from disconnected clusters to conformal maps and real-world data. We demonstrate the use of QCML and quantum geometry in obtaining a geometric understanding of data. In the discussion section (Section~\ref{sec:discussion}), we reaffirm that QCML is intimately connected with quantum geometry and provides new insights into data structure that may be overlooked by classical data analysis methods. Some of the more technical information and details are provided in the Supplementary Information section of the article.

\section{Overview: From Data to Quantum Geometry}
 \label{sec:picture}

Before delving into the technical details of QCML and quantum geometry, we provide a high-level overview of how this work fits within the broader context of geometric data analysis and manifold approximation methods.

Given a smooth manifold or continuous geometric object, it is often important to approximate it using a finite set of real numbers. This is traditionally achieved through triangulations or other graph-based discretization techniques that produce a finite weighted graph. The resulting graph representation can then be used, for example, in numerical geometry or theoretical analysis.
A more modern, though less widely known, approach is the approximation of continuous geometries by \emph{fuzzy} or \emph{matrix} geometries, in which the geometry is encoded in a finite-dimensional noncommutative algebra of Hermitian matrices. This representation, developed in mathematical physics and noncommutative geometry, forms the main foundation of this work. In this approach, one speaks of ``quantizing the geometry,'' where the manifold is viewed as a collection of quantum cells with fixed volume but variable shape, offering a coarse-grained view on the underlying space.

\begin{figure}[htbp]
\centering
\includegraphics[width=\textwidth,height=308.30931000pt]{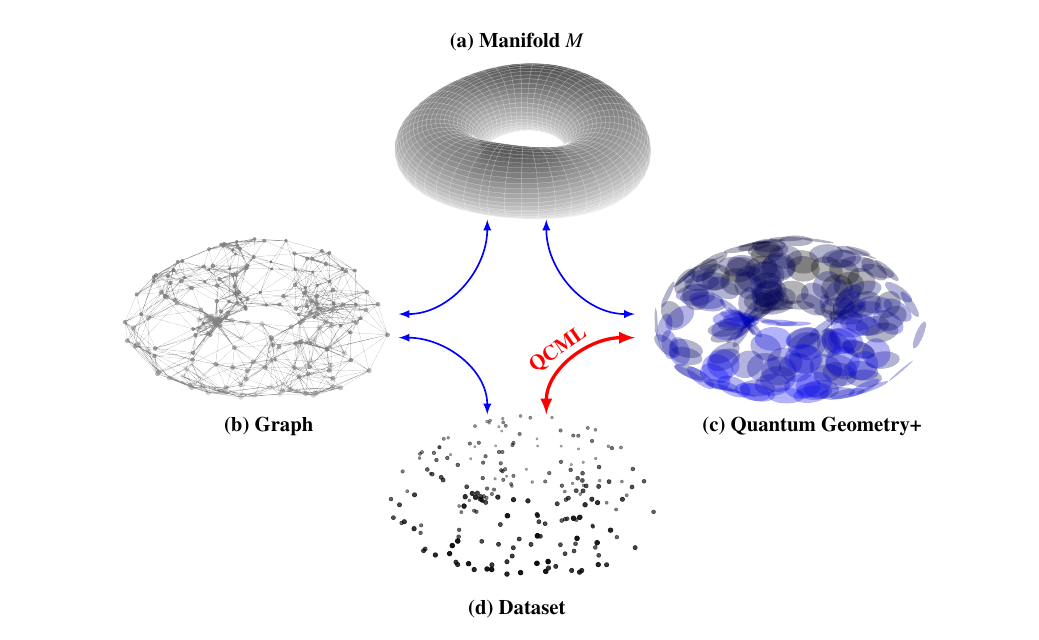}
\caption{Conceptual map placing this work within the context of geometric data analysis and manifold approximations. The top part of the diagram illustrates two established approaches to approximating a smooth manifold~(a): via classical graph-based discretizations~(b), such as triangulations, or via quantum or fuzzy geometries~(c), as developed in mathematical physics. The bottom part of the diagram represents the inverse problem—recovering geometry from data. A dataset~(d), often sampled from a low-dimensional manifold, can be analyzed using graph-based manifold learning methods~(b) (bottom-left blue arrow), or alternatively, via the recently introduced QCML approach~(c) (bottom-right red arrow). QCML constructs a quantum geometry from data using learned observables and quasi-coherent states. This quantum geometric approach is the central focus of this work.}
\label{fig:big-picture}
\end{figure}

These two approaches to manifold approximation---graph-based and matrix geometry-based---are complementary (or dual) to each other and are schematically illustrated in the top part of Figure~\ref{fig:big-picture}.
In contrast, the goal of geometric data analysis is the reverse: it seeks to uncover underlying geometric or manifold structures from finite datasets and to extract meaningful geometric properties from them. This approach is particularly useful in settings where concentration of measure occurs---that is, when data points are clustered near a low-dimensional manifold.

Known techniques such as Laplacian eigenmaps, diffusion maps, and topological data analysis attempt to reconstruct the manifold by building a graph-based approximation from the data.

Only recently, with the introduction of QCML \cite{musaelian2024qcml,candelori2025robust}, has a method appeared that directly constructs \emph{quantum geometry} from data and applies quantum geometric techniques to understand it. In QCML, each data point is lifted to a quantum state in a Hilbert space (a quasi-coherent state), and data features are encoded in a learned set of Hermitian matrices (observables). These matrices and states capture both local and global properties of the data—revealing topological and geometric patterns that may be overlooked by graph-based approaches.

The bottom part of Figure~\ref{fig:big-picture} illustrates the two approaches to data analysis: graph-based methods and quantum geometry. The new connection introduced by QCML, highlighted in red, is the central focus of this work. We examine examples where quantum geometries are learned from both synthetic and real-world datasets and demonstrate how quantum geometric tools can be used to extract intrinsic properties of the data.

\section{Quantum Cognition Machine Learning}
 \label{sec:QCML}

Quantum cognition machine learning (QCML) is a recently developed framework for representing datasets using quantum principles that encode both local and global properties of the data \cite{musaelian2024qcml,candelori2025robust}. { The cognition part of the name reflects the motivation of the method:
to construct compact, structured, and interpretable internal representations of
data, in analogy with cognitive organization of information for reasoning rather than memorization \cite{BusemeyerBruza2025Book}. We use this analogy as motivation and do not claim here to model cognitive mechanisms.}

Here we introduce some basic concepts of QCML, following Ref.~\citeonline{candelori2025robust}. Consider a $T \times D$ dataset $\mathcal{X}$, consisting of $T$ observations on $D$ features or variables. Each data point $x^t$ consists of a $D$-dimensional real-valued vector of data features $x^t=(x_1^t,...,x_D^t) \in \mathbb{R}^D$. 
Given a dataset $\mathcal{X} = \{x^1, \ldots, x^t, \ldots, x^T\} $, QCML associates to each data point $x^t \in \mathcal{X}$ a quantum state $\ket{x^t}$ in a $N-$ dimensional Hilbert space. This is an abstract state which contains  information that can be used to evaluate all of the $D$ features. The state is found as a ground state (state with lowest eigenvalue) of a data-dependent \emph{displacement Hamiltonian} described below. 

The central object in QCML is a \emph{matrix configuration} $X = \{X_1, \ldots, X_a, \ldots, X_D\}$, where each $X_a$ is an $N \times N$ Hermitian matrix. A matrix $X_a$ is what we call an observable that corresponds to the $a^{th}$ variable or feature. { Within the motivating cognitive analogy, \(X_a\) may be viewed as an abstract
concept used to evaluate information encoded in the state \(\ket{x^t}\).} This configuration acts as a non-commutative surrogate for the coordinate functions of an embedding of the data manifold into $\mathbb{R}^D$.   

For each data point $x \in \mathbb{R}^D$, one defines the displacement (or error) Hamiltonian \cite{steinacker2021quantum,ishiki2015matrix}
\begin{equation}
\label{disp-Hamiltonian-0}
    H(x) = \frac{1}{2} \sum_{a=1}^D (X_{a} - x_{a} I_N)^2,
\end{equation}
and takes $\ket{x}$ to be its ground state (state with lowest eigenvalue). 
The matrix configuration assigns to $\ket{x}$ a ``position'' in $\mathbb{R}^D$ via the expectation values
\[
    X(x) = \left( \expval{X_1}{x}, \ldots, \expval{X_D}{x} \right),
\]
which forms the QCML point cloud
\[
 \label{QCML-point-cloud}
    \mathcal{X}_X = \{ X(x^t) \mid x^t \in \mathcal{X} \}.
\]
A deviation of the cloud point $X(x^t)$ from the original point $x^t$ can be measured by the squared \emph{displacement}
\begin{equation}
\label{bias}
    d^2(x) =\left\| X(x) - x \right\|^2 = \sum_a \Big(X_a(x)-x_a \Big)^2. 
\end{equation}
We also characterize quantum fluctuations by the \emph{variance} $\sigma^2(x)$, defined as
\begin{equation}
\label{variance}
    \sigma^2(x) = \sum_a \sigma_a^2(x), \quad \sigma_a^2(x) = \expval{X_a^2}{x} -\expval{X_a}{x}^2.
\end{equation}

The matrix configuration is optimized to approximate the dataset by minimizing a combination of the mean-squared deviation between original data points and their quantum geometric images and by controlling quantum fluctuations. Namely, the loss function is given by
\begin{align}
    L[X] &= \sum_{x \in \mathcal{X}} \left( d^2(x) + w \cdot \sigma^2(x) \right),
 \label{eq:qcmlloss}
\end{align}
so that
\begin{align}
    X_a = \argmin_{X_a \in \mathrm{Mat}(N)} L[X],
 \label{eq:Xkopt}
\end{align}
where $\mathrm{Mat}(N)$ is the space of $N \times N$ Hermitian matrices. Quantum fluctuations, quantified by the variances of $X_a$ on $\ket{x}$, are bounded and weighted by a hyperparameter $w > 0$.  Larger values of $w$ will force the coherent states to be more localized, while lower values ensure that the QCML point cloud is closer to the original data.  For the particular choice $w = 1$, the loss function in \eqref{eq:qcmlloss} coincides with twice the sum of ground state energies $\lambda(x)$ of $H(x)$ taken over $x \in \mathcal{X}$, which can be viewed as a natural choice for $w$. However in practice this value of $w$ is often too high and it may result in matrix configurations that are commutative, with zero quantum fluctuations. These matrix configurations are entirely classical, and the QCML point cloud in this case consists of a $N$-means clustering approximation to the data. The quantum geometry in these cases is entirely lost and therefore these configurations must be avoided. Typically values of $w$ that are closer to zero seem to lead to interesting quantum geometries, but there is no clear principle at the moment for choosing $w$ other than experimentation with the data. { In this sense, \(w\) may be regarded as a resolution or coarse-graining
parameter rather than as a parameter with a unique universally optimal value.}

The result of this optimization is a quantum geometric embedding of the dataset, where the geometry is encoded in the matrix configuration $X$, and structure is revealed by the collective properties of the quasi-coherent states $\ket{x}$. This quantum representation has proven robust to noise and has already been used to extract intrinsic dimensions of datasets in Ref.~\citeonline{candelori2025robust}.

\paragraph{Example 1: Fuzzy sphere $S^2_N$ from random points on a sphere.} As a simple example, consider a dataset consisting of 1000 points distributed uniformly over the surface of a unit sphere $S^2$ embedded in $\mathbb{R}^3$. For this highly symmetric distribution it is expected that the optimal quantum geometry is given by a fuzzy sphere (see Appendix \ref{app:QG-basic-examples}
and Ref. \citeonline{madore1992fuzzy}), namely $X_a =\alpha J_a$, where $J_a$ are angular momenta operators corresponding to the angular momentum $j=\frac{N-1}{2}$ and $\alpha$ is a normalization constant. For the loss function \eqref{eq:qcmlloss} it is easy to find $\alpha = 1/(j+w)$ with the minimal loss function per data point given by $w/(j+w)$.

\begin{figure}[htbp]
\centering
\includegraphics[width=\textwidth,height=200.77798000pt]{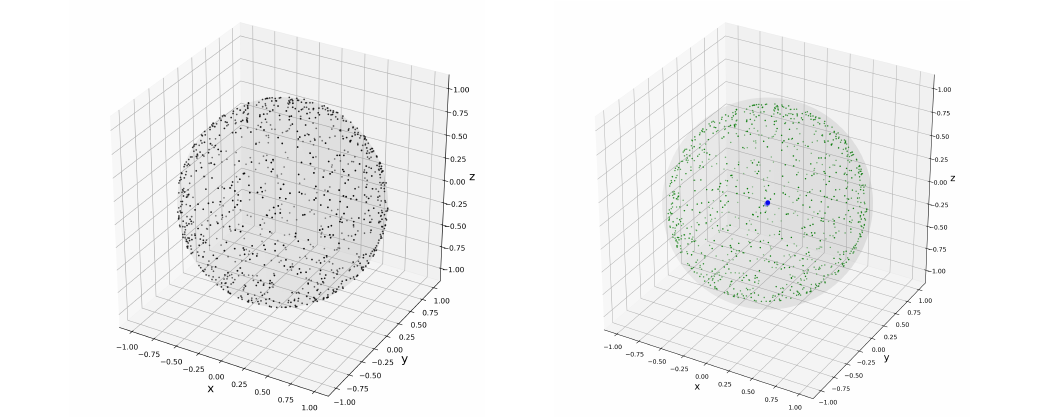}
\caption{Learned geometric representations from trained quantum operators \(\mathbf{X}\). The unit sphere is shown for reference. Left: The original data, consisting of 1000 points generated on the surface of a unit sphere with a uniform measure, is shown in black. Right: Quantum-learned QCML point cloud (green) with degeneracy points (blue). There are three nearby degeneracy points, each with topological charge $-1$, separated by a distance of about 0.03. The training was performed with hyperparameters \(N=4\), \(w=0.1\). }
\label{fig:1-sphere-N4-E20000}
\end{figure}

As an example we present the QCML learned operators \( X_{1,2,3} \). We used $N=4$ so that the operators are represented by Hermitian \( 4 \times 4 \) matrices and $w=0.1$ for the quantum fluctuation weight. To show that the learned operators, normalized as $J_a = -(j + w) X_a$, are close to angular momentum generators for $j=3/2$, we compute:
\begin{align}
\mbox{eigenvalues of }J_3  \approx \{-1.50, -0.49, +0.51, +1.52\},
\quad
 \|[J_a,J_b]-i\epsilon_{abc}J_c\| \approx 0.16, 
 \quad
 \|\sum_a J_a^2 - j(j+1)\| \approx 0.11, 
 \label{learned-X-100}
\end{align}
{ where \(\|\cdot\|\) denotes the Hilbert--Schmidt norm defined in
Eq.~\eqref{inner-Mat}.}
Indeed, we see that the eigenvalues of the operator \( J_3 \) are close to those of the angular momentum generator. The small norms confirm both the commutation relations and the value of the quadratic Casimir $J_a^2$ for the  $j=3/2$ representation. Accuracy can be further improved with longer training.
We expect the displacement Hamiltonian to exhibit degeneracy points near the origin, where the ground state becomes degenerate. These points are marked in blue in the figure. { Since the learned operators are not exact angular-momentum generators, the rotational symmetry is weakly broken. Consequently, the single degeneracy of the exact $N=4$ fuzzy sphere, carrying total monopole charge $-\left(N-1\right)=-3$, splits into three closely spaced degeneracy points, each with topological charge $-1$.}

The outcome of the QCML learning process is a representation of the dataset as a collection $X$ of Hermitian matrices and a set of quasi-coherent states $\{\ket{x^t}\}$, with the number of matrices $D$ equal to the number of features in the dataset. This collection serves as a model of the dataset and can be visualized through the ground states of the corresponding displacement Hamiltonians. The training procedure favors configurations in which the learned operators $X$ are as close to mutually commuting as possible, given the constraints imposed by the Hilbert space dimension and dataset size. Such matrix configurations have been extensively studied in the context of quantum or matrix geometry. We briefly review the basic concepts of quantum geometry in Section ~\ref{sec:QG} and provide additional technical details in Appendices ~\ref{app:matrix-laplacian} and ~\ref{sec:structural}.

\begin{figure}[htbp]
\centering
\includegraphics[width=\textwidth,height=196.68080000pt]{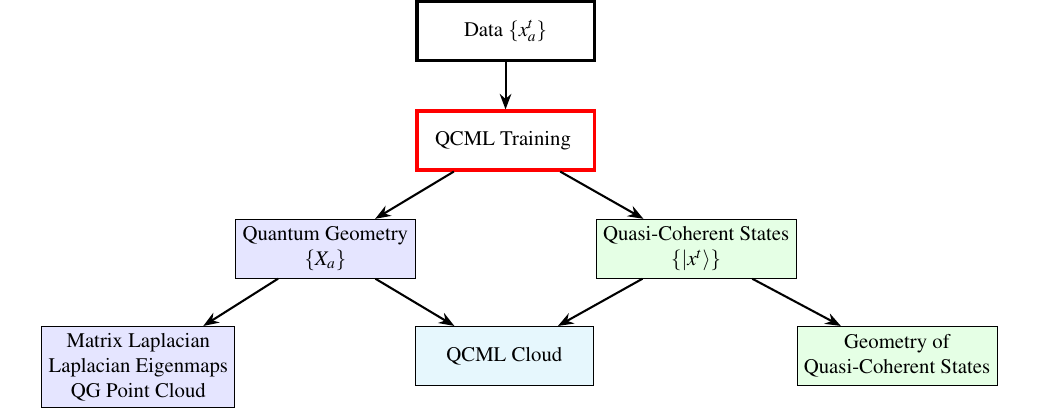}
\caption{Schematic diagram illustrating QCML procedure. Observables $X_a$ (quantum geometry) and quasi-coherent states $\ket{x^t}$ are obtained through the training using data dependent displacement Hamiltonians and the loss function optimizing a combination of displacement and variance of the data. The learned operators and states are then used to obtain quantum-geometric characteristics of the data. Then these characteristics can be used for extracting intrinsic dimension, clustering, topological properties, inference (expectation values), etc.}
\label{fig:scheme-QCML}
\end{figure}

We schematically represent the QCML data analysis in the diagram in Figure~\ref{fig:scheme-QCML}. We comment that QCML does not retain the original data points; instead, it outputs a set of observables $\{X_a\}$ and quasi-coherent states $\{\ket{x^t}\}$. The learned observables $\{X_a\}$ define the \emph{quantum geometry}. While the quasi-coherent states are not part of the quantum geometry proper, their structure can be analyzed separately—this is referred to as the \emph{geometry of quasi-coherent states}, which lies beyond the scope of this article (see Section~\ref{sec:discussion} for brief comments). However, in the examples presented in Section~\ref{sec:examples}, we use the concept of the \emph{QCML cloud}, defined as the set of expectation values $\bra{x^t}X_a\ket{x^t}$, which incorporates both observables and quasi-coherent states. Therefore, the QCML data can be understood as an augmented version of quantum geometry.

\section{Quantum geometry: essential introduction}
 \label{sec:QG}

In this Section, we provide a concise overview of quantum geometry, focusing on the key concepts relevant to the examples that follow. It can be skipped or used as a reference. This summary is necessarily brief and does not capture the depth of the subject. For a more comprehensive review, we refer the reader to Ref.~\citeonline{steinacker2021quantum} (Sections 2 and 3) and Ref.~\citeonline{steinacker2024quantum} (Chapters 3 and 4). A glossary of less conventional terms is provided in Appendix~\ref{app:glossary}.

In \emph{fuzzy} or \emph{quantum geometry}, functions on a manifold are replaced by matrices, and geometric structures are encoded through a specific set of matrices \cite{madore1992fuzzy,balachandran2007lectures}. The corresponding matrix algebras serve as finite-dimensional approximations to the algebra of functions on smooth manifolds, capturing essential topological and differential properties. This approach resulted from the advances in noncommutative geometry and matrix models that put forward an idea that classical notions of space and geometry can emerge from a more fundamental quantum substrate—one based on operators and states in a complex Hilbert space. Quantum geometry can also be understood as an analog of a phase space of classic mechanics appearing as a result of the quantization. In particular the size of the Hilbert space $N$ corresponds to
the inverse Planck's constant $1/\hbar$. In the limit $\hbar\to 0$ ($N\to\infty$) the classical geometry is restored. Speaking semi-classically the quantum-mechanical uncertainty relation divides phase space into cells with volume proportional to powers of $\hbar$ and corresponds to a coarse-grained point of view on the phase space geometry. Importantly, this discrete representation of geometry is grid-free. 
This approach has found numerous applications in high-energy physics \cite{kabat1997spherical,zhu2023uncovering,chatzistavrakidis2011intersecting,berenstein2012matrix}. In particular, fuzzy spaces have been used to regularize quantum field theories, giving rise to noncommutative gauge theories and matrix models of emergent spacetime \cite{steinacker2010emergent,steinacker2020quantum,steinacker2024quantum}.

These ideas have long played a central role in mathematical and theoretical physics, now with QCML they are being adapted to data analysis.

The central idea of quantum geometry is to encode a geometric subspace \( \mathcal{M} \hookrightarrow \mathbb{R}^D \) of the feature space \( \mathbb{R}^D \) using a set of matrices—and, conversely, to interpret certain matrix configurations as approximating such subspaces. More precisely, a \textbf{matrix configuration} consists of \( D \) Hermitian \( N \times N \) matrices \( \{X_a\}_{a=1}^D \), often referred to as \emph{observables}. The space \( \mathrm{Mat}(N) \) of Hermitian \( N \times N \) matrices is equipped with an inner product and norm given by
\begin{align}
\label{inner-Mat}
    \langle \Phi, \Psi \rangle = \operatorname{Tr}(\Phi^\dagger \Psi), \qquad 
    \|\Phi\|_2^2 = \langle \Phi, \Phi \rangle \,.
\end{align}
This structure makes \( \mathrm{Mat}(N) \) a Hilbert space and, moreover, a Banach algebra, since \( \|\Phi \Psi\| \leq \|\Phi\| \|\Psi\| \).

By analogy with quantum mechanics, we seek to identify the space of observables \( \mathrm{Mat}(N) \) with the space \( L^2(\mathcal{M}) \) of square-integrable functions on a \( d \)-dimensional manifold \( \mathcal{M} \), equipped with the inner product
\begin{align}
\label{inner-M}
    \langle \phi, \psi \rangle = \int\limits_{\mathcal{M}} \Omega\, \phi^* \psi, \qquad \quad \|\phi\|_2^2 = \int\limits_{\mathcal{M}} \Omega\, |\phi|^2
\end{align}
and the associated norm.
Here \( \Omega \) denotes the underlying (symplectic) measure on \( \mathcal{M} \).
This leads to a correspondence
\begin{align}
\label{class-Quant-rel}
    \mathrm{Mat}(N) \ \stackrel{\sim}{\longleftrightarrow} \ L^2(\mathcal{M}),
\end{align}
which should preserve the inner products \eqref{inner-Mat} and \eqref{inner-M} at least in an appropriate ``semiclassical'' regime, along with additional structure discussed below. In particular, the observables \( X_a \) in a given matrix configuration are interpreted as quantized embedding functions of \( \mathcal{M} \) into the feature space:
\begin{align}
\label{embedding-map}
    X_a \sim x_a: \ \mathcal{M} \to \mathbb{R}^D.
\end{align}
This idea proves remarkably effective for encoding large datasets and can be implemented in practice as follows.

The key tool for establishing the correspondence in \eqref{class-Quant-rel} between a given matrix configuration \( \{X_a\} \) and a ``quantum'' geometry is the \textbf{displacement Hamiltonian} introduced in \eqref{disp-Hamiltonian-0}. This is a positive Hermitian operator associated to each point \( (x_a) \in \mathbb{R}^D \) in feature space. Since \( H(x) \) is positive (semi)definite, we can determine its ground state(s) and corresponding energy. For generic points \( x \in \mathbb{R}^D \), the ground state is expected to be non-degenerate. In such cases, we define the \textbf{quasi-coherent state} \( |x\rangle \) as the unique ground state of \( H(x) \) with the ground state energy $\lambda(x)$ (the minimal eigenvalue):
\begin{align}
    H(x) |x\rangle = \lambda(x) |x\rangle \,.
\end{align}

As in quantum mechanics, these quasi-coherent states can be used to compute the expectation values of observables. The ground state energy satisfies the decomposition
\begin{align}
\label{lambda-identity}
    2\lambda(x) = \sigma^2(x) + d^2(x),
\end{align}
where \( \sigma^2(x) \) is the total quantum uncertainty (\ref{variance}) of the observables \( X_a \) in the state \( |x\rangle \), and
$d^2(x)$
measures the squared displacement between the point \( x \in \mathbb{R}^D \) and the expectation values \( \langle X_a \rangle = \langle x | X_a | x \rangle \) in the quasi-coherent state (\ref{bias}). 
Hence, the map
\begin{align}
\label{targetspace-map}
  \tilde{\mathbb{R}}^D \to \mathbb{R}^D, \qquad  x \mapsto \langle x|X_a|x\rangle
\end{align}
associates to each point \( x \) in feature space the expectation values of the observables \( X_a \), interpreted as the closest and optimally localized point on the emergent manifold \( \mathcal{M} \). Here, \( \tilde{\mathbb{R}}^D \subset \mathbb{R}^D \) denotes the set of points for which the ground state of \( H(x) \) is non-degenerate. This map is highly nonlinear yet smooth and typically defines an approximate projection from \( \tilde{\mathbb{R}}^D \) onto \( \mathcal{M} \). 

The full quantum geometry can be visualized by sampling random points in a region of feature space and plotting their images under the map \eqref{targetspace-map}. The resulting quantum geometry point cloud—or \textbf{QG point cloud}—gives an effective representation of \( \mathcal{M} \). This picture can be further enriched by visualizing additional quantities, such as the uncertainty \( \sigma(x) \) of the quasi-coherent states, using color or size to encode fluctuations.

The quantities \( \lambda \) and \( \sigma \) serve as indicators of the quality of the quantum geometry at a given point \( x \); small values of \( \lambda \) and \( \sigma \) correspond to high fidelity. Generically, such quantum geometries are expected to consist of \( N \) quantum cells, labeled by \( i = 1, \ldots, N \). 
As in quantum mechanics, these quantum cells do not have fixed shape or volume, but they provide an intuitive picture of the information content and intrinsic uncertainty in the framework.

By selecting a representative point \( x_i \in \mathbb{R}^D \) within each quantum cell, we obtain the approximations
\begin{align}
\label{trace-phi}
    \Tr(\Phi) \approx \int\limits_\mathcal{M} \Omega\, \phi(x) \approx \sum_i \phi(x_i), \qquad  \|\Phi\|_2^2 \approx \int\limits_\mathcal{M} \Omega\, |\phi(x)|^2 \approx \sum_i |\phi(x_i)|^2 \approx \|\phi\|_2^2
\end{align}
for matrices \( \Phi \) corresponding to functions \( \phi \) in the semiclassical regime, cf. \eqref{inner-M}. In particular, for the identity matrix \( \one \), this yields an estimate for the dimension \( N \) of the Hilbert space:
\begin{align}
\label{L2-norm-estimate2}
   N = \|\one\|_2^2 \approx \int\limits_\cM \Omega\,  \approx \sum_i 1 \ .
\end{align}

\paragraph{Commuting matrix configurations.}
An important example of degenerate quantum geometry is provided by a matrix configuration with vanishing commutators, \( [X_a, X_b] = 0 \). In this case, there exists a common eigenbasis \( |i\rangle \) such that
\[
X_a = \sum_i x_a^{(i)} |i\rangle\langle i|.
\]
This defines \( N \) points \( x_a^{(i)} \in \mathbb{R}^D \) in feature space. Since the displacement Hamiltonian is diagonal in this basis, the quasi-coherent state \( |x\rangle \) is simply the eigenstate \( |i\rangle \) corresponding to the point closest to \( x \). The ground state energy \( \lambda(x) \) is then half the squared distance from \( x \) to the nearest \( x^{(i)} \).
Thus, for a given dataset, the commuting matrix configuration that minimizes the total ground state energy effectively recovers the $K$-means clustering into \( N \) centroids \( x_a^{(i)} \in \mathbb{R}^D \).

While simple and intuitive, commuting configurations suffer from lattice artifacts and lack smoothness.  In contrast, quantum geometries based on noncommuting \( N \times N \) matrices yield fuzzy but smooth approximations of continuous spaces.

A few  basic examples of genuine quantum geometries discussed in the literature \cite{steinacker2021quantum} are presented in Appendix \ref{app:QG-basic-examples}.

In the context of QCML, the matrix configuration \( \{X_a\} \) is not fixed a priori but learned directly from the data.

\subsection{Quantum geometric structure of \texorpdfstring{$\mathcal{M}$}{M}}
 \label{sec:QGstructure}

Given a matrix configuration \( \{X_a\} \), we define the {\bf abstract quantum space} \( \mathcal{M} \) to be the set of quasi-coherent states, modulo phase\cite{steinacker2021quantum}:
\begin{align}
\label{abstract-quantum-space}
    \mathcal{M} := \{ |x\rangle \;|\; x \in Q \}/_{U(1)} \quad \subset \mathbb{C}P^{N-1},
\end{align}
where \( Q \subset \mathbb{R}^D \) is a sufficiently large hypercube in feature space containing the relevant data.

The space \( \mathcal{M} \) can be embedded into feature space via the map
\begin{align}
\label{embedding-map2}
     \mathcal{M} \to \mathbb{R}^D, \qquad |x\rangle \mapsto \langle x|X_k|x\rangle,
\end{align}
which refines the projection \eqref{targetspace-map}. This embedding can be visualized by sampling random points in \( Q \); their images under \eqref{embedding-map2} typically concentrate around a submanifold of \( \mathbb{R}^D \), which can be interpreted as an approximation to the underlying data manifold.\footnote{This often involves a mild “thickening” or oxidation of the data manifold (cf.~Ref.~\citeonline{felder2024oxidation}), which is not problematic in the present context.} 

Viewing \( \mathcal{M} \) as a subspace of the Hilbert space—or more precisely, of projective Hilbert space \( \mathbb{C}P^{N-1} \)—reveals additional geometric and algebraic structure, justifying the designation of \( \mathcal{M} \) as a ``quantum space.''

As in quantum mechanics, the embedding into complex Hilbert space induces natural geometric structures on \( \mathcal{M} \): a connection \( A \) (also known as Berry connection in the context of adiabatic theory), a closed two-form \( \omega = dA \), and a \emph{quantum metric} \( g \). These are obtained as pull-backs of the symplectic structure and the Fubini–Study metric on \( \mathbb{C}P^{N-1} \). Together, they form the \emph{quantum geometric tensor} \( q_{\mu\nu} \), which is encoded in the inner product of quasi-coherent states, $
\langle x|y\rangle = \exp\{i\varphi(x,y) - D^2(x,y)\}$,
and can be computed explicitly as
\begin{align}
\label{A-omega-def}
    q_{\mu\nu} = \tfrac{1}{2}(g_{\mu\nu} + i \omega_{\mu\nu}) 
    = (\partial_\mu + i A_\mu)\langle x|\;(\partial_\nu - i A_\nu)|x\rangle, 
    \qquad i A_\mu = \langle x|\partial_\mu|x\rangle.
\end{align}

These structures are defined entirely within the quantum framework, without requiring a semiclassical approximation, and can be readily computed numerically. They are useful, for instance, in estimating the intrinsic dimension of \( \mathcal{M} \) \cite{candelori2025robust}. Importantly, the function \( D(x, y) \) measures the intrinsic distance between two points \( x, y \in \mathcal{M} \) with respect to the quantum metric \( g \), rather than their distance in feature space.

This additional structure on \( \mathcal{M} \) offers qualitatively new perspectives for data representation that go beyond classical methods. For instance, the distance on \( \mathcal{M} \) defined by the quantum metric \( g \) captures the intrinsic proximity between states \( |x\rangle \), which may differ significantly from distances in feature space. 
Most strikingly, distinct points on \( \mathcal{M} \) can be mapped to the same point in feature space via \eqref{embedding-map}, even if they are widely separated with respect to the quantum metric. This enables the coherent modeling of different objects that share the same classical features—for example, a single word or phrase with multiple, unrelated meanings.
The symplectic structure \( \omega \) quantifies the local intrinsic uncertainty of \( \mathcal{M} \) and can be computed via the parallel transport induced by the connection \( A \). Moreover, by using suitable variants of the displacement Hamiltonian, one can define non-local quantum states that reveal non-classical correlations, similar to those familiar from quantum mechanics.
All of this emerges naturally from the complex structure of Hilbert space and operators underlying the present framework. A practical use case for the state proximity is discussed in Ref.~\citeonline{rosaler2025supervised}.

{ We emphasize that matrix quantum geometry is not restricted to even-dimensional spaces. Although a nondegenerate symplectic structure is necessarily even-dimensional, matrix configurations may also describe
degenerate Poisson geometries and odd-dimensional manifolds. A simple example is provided by
\[
X_1=S_x/s,
\qquad
X_2=S_y/s,
\]
where \(S_x\) and \(S_y\) are the spin-\(s\) generators. The corresponding displacement Hamiltonian is
\[
H(x,y)
=
\frac12\left[
\left(x-S_x/s\right)^2
+
\left(y-S_y/s\right)^2
\right].
\]
Writing \(x=r\cos\phi\) and \(y=r\sin\phi\), rotations in the \((x,y)\)-plane are implemented by unitary transformations generated by \(S_z\). Consequently, the lowest eigenvalue of \(H(x,y)\) depends only on
\(r\), and its ground states are localized on a circle of radius \(r_*(s)\). For example,
\[
r_*\!\left(1/2\right)=1,
\qquad
r_*(1)=\sqrt{15}/4,
\]
and \(r_*(s)\to 1\) as \(s\to\infty\). Thus the pair
\((S_x/s,S_y/s)\) provides a noncommutative finite-dimensional realization of the odd-dimensional geometry \(S^1\).
}

The numerical dimension estimates below use the response matrix
\(R_{ab}(x)=\partial\langle x|X_a|x\rangle/\partial x_b\), with the learned
operators held fixed. Its spectral representation contains inverse energy
gaps, whereas the quantum metric of Eq.~\eqref{A-omega-def}, expressed in
the same coordinates, contains inverse squared gaps. For a nondegenerate ground state,
the two matrices have the same zero-eigenvalue subspace, but their nonzero
eigenvalues may differ. The threshold
\(0.5\) used below applies to the response eigenvalues.

\subsection{Topological properties of \texorpdfstring{$\mathcal{M}$}{M}}
\label{sec:topological}

The framework of quantum geometry allows not only for the identification of disconnected components of \( \mathcal{M} \) (see Section~\ref{sec:eigenmaps}), but also for the extraction of finer topological features. We illustrate this through the second cohomology group \( H^2(\mathcal{M}) \), as encoded in the cohomology class of the closed two-form \( \omega \). 
For any two-cycle \( S^2 \subset \mathcal{M} \), one 
can define integer-valued topological charges via
\begin{align}
\label{chern-number}
   c_1 :=  \int_{S^2} \frac{\omega}{2\pi}  \ \in \mathbb{Z},
\end{align}
known as the first Chern number. The integral (\ref{chern-number}) can be interpreted as ``magnetic'' flux through $S^2$, which is always integer-valued because \( \omega \) is the curvature of the connection one-form \( A \) defined in \eqref{A-omega-def} on the line bundle of quasi-coherent states over \( \mathcal{M} \). In the context of quantum mechanics, \(\omega \) is known as Berry curvature. 
Any non-zero $c_1 \neq 0$ implies that the embedding $S^2 \subset \cM$ is non-contractible, reflecting a non-trivial topological structure of $\cM$.
In practice, we compute such integrals over spheres \( S^2 \subset \mathbb{R}^3 \subset \mathbb{R}^D \), chosen to encircle isolated degeneracy points of the displacement Hamiltonian \( H(x) \) { in some suitable plane $\R^3$ in feature space, or possibly in the feature space defined by some of the lowest Laplacian eigenmaps.} These are points in feature space where the ground state becomes degenerate. Since \( \omega \) is both smooth and quantized, the integral \eqref{chern-number} can be nonzero only if \( S^2 \) encloses such degeneracies. These degeneracy points are typically isolated in $\mathbb{R}^3$ and can be detected numerically. Commonly referred to as ``monopoles,'' they act as sources of Berry curvature and are characterized by integer topological charges \eqref{chern-number}.
{ Numerically, an isolated degeneracy is enclosed by a small triangulated sphere
in a three-dimensional subspace of feature space, and the Berry flux is
computed from the ground states on this sphere. The sphere radius and mesh
resolution are varied to verify that the resulting flux is stable and close to
an integer. We report an integer charge only when this stability criterion is
satisfied; separate statistical error bars are not assigned.}

In the right panel of Figure~\ref{fig:1-sphere-N4-E20000} we showed in blue the degeneracy point found numerically for the uniform sphere example (fuzzy $S^2_4$). { For the ideal fuzzy-sphere geometry there is a single degeneracy point at the
origin with topological charge \(-3\). For the QCML-learned matrices, this
splits into three nearby degeneracy points, each with charge \(-1\).}

Chern numbers and monopoles offer powerful geometric and topological tools for understanding the structure of datasets, especially those represented on manifolds or exhibiting internal symmetries. Chern classes capture global properties of vector bundles over data manifolds, identifying when it is impossible to define consistent, global coordinates or features across the entire dataset, revealing intrinsic curvature or twisting in the underlying structure. Monopoles represent topological defects or singularities, often reflecting global phase behavior or nontrivial connections in the data. This opens the possibility of classifying datasets based on topological signatures, yielding robust, interpretable features that are stable under deformation or noise. Quantum geometry tools enable the systematic extraction of topological invariants from matrix configurations. Generalizations to higher cohomology groups and characteristic classes are possible; see Ref.~\citeonline{steinacker2024quantum}, Sec.~4.2.1.

\subsection{The matrix Laplacian}
\label{sec:matrix-Laplacian}

If the intrinsic quantum structure is sufficiently fine, the quantum geometry can be approximated by a {\em semi-classical} geometry, relating observables to classical functions as in \eqref{class-Quant-rel}. In this regime, commutators of observables reduce to \( \sqrt{-1} \) times the Poisson brackets of the corresponding classical functions, consistent with the correspondence principle of quantum mechanics. The two-form \( \omega \) can then be interpreted as the symplectic form associated with the Poisson structure on \( \mathcal{M} \). In particular, large values of \( |\omega| \) correspond to small local uncertainty.
Moreover, the commutators
\begin{align}
\label{nabla-def}
\nabla_a := [X_a, \, \cdot\, ] \sim i \{ x_a, \, \cdot\, \}
\end{align}
can be viewed as quantized Hamiltonian vector fields on \( \mathcal{M} \), generated by the quantized functions \( X_a \sim x_a: \mathcal{M} \to \mathbb{R}^D \) that define the embedding map \eqref{embedding-map2}. This perspective provides a geometric understanding of structures such as the matrix Laplacian.

The matrix Laplacian plays a role analogous to the Laplace–Beltrami operator on a classical manifold and is central to understanding the global geometry of the learned quantum space.

Given any matrix configuration \( \{X_a\} \), we define the {\bf matrix Laplacian} as
\begin{align}
\label{eq:matrixLaplacian}
    \Delta = \sum\limits_a [X_a, [X_a, \cdot]].
\end{align}
This construction is equivalent to $\Delta = \nabla_a\nabla_a$, where  $\nabla_a$ is defined in equation~(\ref{nabla-def}).
The matrix Laplacian is a Hermitian, positive-semidefinite operator acting on the space \( \mathrm{Mat}(N) \) of matrices. Its spectrum and eigenmatrices—referred to as \emph{eigenmaps}—encode meaningful geometric features of the quantum space \( \mathcal{M} \):
\begin{align}
    \Delta Y_i = \lambda_i Y_i, \qquad \mathrm{spec}(\Delta) = \{\lambda_0, \lambda_1, \ldots, \lambda_{\mathrm{max}}\}.
\end{align}

The spectrum reflects properties such as connectedness and effective dimensionality of the quantum geometry. In particular, zero modes correspond to sectors of the matrix configuration that are mutually commuting or disconnected. For irreducible configurations, the asymptotic distribution of eigenvalues can be used to estimate the intrinsic dimension of \( \mathcal{M} \) via Weyl's law:
\begin{align}
    N(\lambda) \sim \lambda^{d/2} \qquad \text{as } \lambda \to \infty,
\end{align}
where \( N(\lambda) \) denotes the number of eigenvalues less than or equal to \( \lambda \), and \( d \) is the intrinsic dimension.

{ The fitting window for extracting the spectral dimension from the eigenvalues $\{\lambda_1,..., \lambda_{\rm max}\}$ should be centered around the geometric mean $\lambda_{\rm NC} = \sqrt{\lambda_1 \lambda_{\rm max}}$ of the lowest and highest eigenvalues of $\Delta$, which characterizes the scale of noncommutativity of the quantum geometry. This arises from the uncertainty relation $L_{\rm UV} L_{\rm IR} \approx L_{\rm NC}^2$ for optimally localized states. The lowest (non-trivial) eigenvalues depend on global features of the manifold, while the highest eigenvalues are far above the semi-classical regime and should be excluded.
  In any case, this spectral dimension is a global observable which provides complementary information to the more detailed local dimension extracted from the quasi-coherent states \cite{candelori2025robust}.}

\begin{figure}[ht]
\centering
\includegraphics[width=\textwidth,height=186.40707000pt]{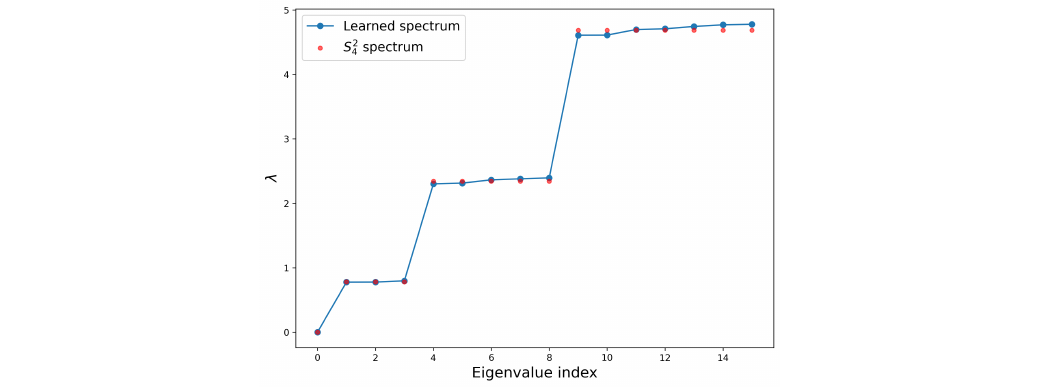}
\caption{Comparison between the spectrum of the matrix Laplacian learned from the data shown in Figure~\ref{fig:1-sphere-N4-E20000} and the exact spectrum of the fuzzy sphere $S^2_4$. The blue points correspond to the spectrum learned by the model, while the red points show the rescaled exact eigenvalues $\ell(\ell+1)/(j+w)^2$ with multiplicities $2\ell+1$, where $\ell = 0, 1, \dots, 2j=3$.}
\label{fig:laplacian-spectrum-fuzzy-sphere}
\end{figure}

As a simple illustration of the use of matrix Laplacian spectrum let us go back to the example of a uniform spherical distribution in Figure~\ref{fig:1-sphere-N4-E20000}. We construct the matrix Laplacian from the learned $X$-matrices (\ref{learned-X-100}) and present its spectrum in Figure~\ref{fig:laplacian-spectrum-fuzzy-sphere}. For the fuzzy sphere corresponding to angular momenta in the $N=2j+1$ dimensional representation the spectrum is given by exact eigenvalues $\ell(\ell+1)$ with degeneracies $2\ell+1$ and $\ell=0,1,\ldots,2j$. We can see from the figure that the Laplacian spectrum corresponding to the QCML learned matrices for $N=4$ (\ref{learned-X-100}) matches after suitable rescaling the exact values for the fuzzy sphere. 

More detailed information on the spectral and variational properties of \( \Delta \) is provided in Appendix~\ref{app:matrix-laplacian}.

In situations where the ambient feature space dimension \( D \) is much larger than the intrinsic dimension \( d \) of the data, it becomes desirable to capture the underlying data manifold using a reduced set of matrices. A powerful approach to this problem is described in the next section.

\subsection{Laplacian eigenmaps and reduced matrix configurations}
\label{sec:eigenmaps}

In classical statistics, Principal Component Analysis (PCA) is based on the covariance matrix \( \Sigma \) of a random variable. The goal of PCA is to capture the dominant modes of variation in the data by computing the eigenvectors of an empirical estimate of \( \Sigma \). These eigenvectors identify the directions of maximal variance, while the corresponding eigenvalues quantify the extent of variability along each direction.

In graph-based learning, the graph Laplacian \( L \) is defined as the difference between the degree matrix and the adjacency matrix. It encodes the connectivity of the graph and provides a notion of smoothness for functions defined on its nodes. The eigenvectors of \( L \) with small eigenvalues correspond to functions that vary most smoothly across adjacent nodes. Techniques such as Laplacian eigenmaps and diffusion maps exploit this structure: the leading eigenvectors represent the slowest decaying modes of diffusion and reveal the large-scale geometry of the underlying manifold.

In quantum geometry, the matrix Laplacian defined in \eqref{eq:matrixLaplacian} plays a role analogous to that of the classical Laplacian, acting as a second-order differential operator that encodes curvature and diffusion properties—now within a noncommutative algebraic structure. Such operator-based approaches have been extensively developed in the study of quantum manifolds, particularly in the work of Steinacker \cite{steinacker2021quantum,steinacker2024quantum}.

The connection between the Laplacian and commutator structures is particularly profound. In classical settings, the Laplacian measures local variation through second derivatives; in the quantum setting, double commutators provide a natural noncommutative analogue, capturing notions of curvature and diffusion. The matrix Laplacian thus serves as a quantum analog of the classical Laplacian, unifying diffusion, uncertainty, and geometry within a single operator framework. This observation motivates the development of dimensionality reduction methods based on the spectral properties of the matrix Laplacian.

To understand qualitative features of the quantum geometry, we propose replacing the original high-dimensional model with a reduced model defined on a lower-dimensional abstract feature space. Specifically, we seek a set of matrices \( \{Y_i\} \) that provide an optimally flat representation of the matrix geometry, in the spirit of Ref.~\citeonline{belkin2003laplacian}. More precisely we look for a reduced set of $n$ Hermitian matrices maximally compatible with the quantum geometry given by $X_a$, that is we aim to minimize $\sum_{i=1}^n\sum_{a=1}^D \tr [X_a,Y_i]^\dagger [X_a,Y_i]$ subject to normalization condition $\tr Y_i^2 =1$. 

This is achieved by selecting the first \( n \) eigenmaps of the matrix Laplacian,
\begin{align}
\Delta Y_i = \lambda_i Y_i, \qquad i = 1, \ldots, n,
 \label{eigenmaps-100}
\end{align}
and interpreting them as a {\bf reduced (abstract) matrix configuration} \( \{Y_i\} \). This yields an abstract representation of the underlying quantum geometry, viewed as an optimally flat embedding \( Y: \mathcal{M} \hookrightarrow \mathbb{R}^n \).

To faithfully capture the geometry of \( \mathcal{M} \), the number of components \( n \) should be chosen to match or exceed its intrinsic dimension. We  require the matrices \( Y_i \) to form an orthonormal basis, 
\[
\mathrm{Tr}(Y_i^\dagger Y_j) = \delta_{ij}, \qquad \text{in particular} \quad \|Y_i\|_2 = 1.
\]

By construction, the embedding defined by the matrices \( Y_i \) has minimal geometric energy or curvature (see Appendix~\ref{app:matrix-laplacian}); that is, it is as flat as possible, in analogy to Ref.~\citeonline{belkin2003laplacian}. This also implies that the relative uncertainty of each \( Y_i \) is minimized. As such, the set \( \{Y_i\} \) provides an optimal semi-classical representation \( \mathcal{M}_Y \) of \( \mathcal{M} \), particularly well suited for extracting the structure of the quantum geometry—most notably its topology.

Moreover, the intrinsic quantum space \( \mathcal{M}_Y \subset \mathbb{C}P^{N-1} \), defined via the quasi-coherent states of the reduced matrix configuration, is expected to closely approximate that of the original. Accordingly, geometric quantities such as the quantum metric \( g \) and symplectic form \( \omega \) are expected to be similar.

The reduced matrix configuration \( \{Y_i\} \) can therefore be regarded as an abstract model for the intrinsic quantum geometry. This model can be visualized by constructing the embedding map~\eqref{embedding-map2}, using the displacement Hamiltonian built from the matrices \( Y_i \) as defined in~\eqref{disp-Hamiltonian-0}.

In particular, we define the \textbf{reduced matrix Laplacian} as
\begin{align}
    \tilde{\Delta} := \sum_{i=0}^n [Y_i, [Y_i, \cdot]],
 \label{redLapl}
\end{align}
for a suitable choice of \( n \). { The mode \(Y_0\) is proportional to the identity matrix and therefore does not contribute to the reduced matrix Laplacian \eqref{redLapl}, since \([A,Y_0]=0\) for any matrix \(A\).} This reduced operator is useful for extracting structural features of the learned model, such as intrinsic dimension, topological characteristics, and other potentially hidden properties of the data. However, it captures the geometry of the original feature space only indirectly.

One may carry over the data  by
mapping the original quasi-coherent states \( |x_j\rangle \) corresponding to the data points into the abstract feature space via the expectations \( \langle x_j | Y_i | x_j \rangle \). This procedure transforms raw features into more conceptual representations—for example, converting unstructured image pixels into perceptually meaningful features.

A more elegant method for obtaining a smoothed quantum geometry, while staying within the original feature space, is to orthogonally project the original observables \( X_a \) onto the span of the eigenmaps \( Y_i \):
\begin{align}
    \tilde{X}_a := \sum_i b_{ai} Y_i, \qquad b_{ai} = \mathrm{Tr}(Y_i^\dagger X_a ).
\end{align}
The resulting matrix configuration \( \{ \tilde{X}_a \} \) defines a UV-regularized version of the quantum geometry in the feature space. Since the eigenmaps \( \{Y_i\} \) form an orthonormal basis of \( \mathrm{Mat}(N) \), this procedure effectively filters out high-frequency modes, preserving the essential geometric and metric structure of the data.

In this work we focus on smooth geometric objects and use semiclassical descriptions. 
However, when additional insight into the nature of the observables is available, it may be preferable {\em not} to smooth out certain discontinuous operators—such as classifying observables or ``quantum observables'' of the form \( X = |x\rangle\langle y| + |y\rangle\langle x| \). These operators can encode correlations between distant or disconnected regions of the quantum geometry, highlighting a key strength of the QCML approach. In particular, such ultra quantum observables can be used for supervised QCML training which is out of the scope of this work.

More broadly, the spectra of operators not only encode geometric structure but also offer a universal language in which geometry, physics, and data representations may be seen as emergent from spectral properties alone.

\paragraph{Zero modes and separating connected components of \( \mathcal{M} \).}

For irreducible matrix configurations, the Laplacian \( \Delta \) is strictly positive, with the exception of the trivial zero mode given by the identity matrix \( \one \). This is no longer the case for reducible matrix configurations composed of \( k \) irreducible blocks, such as
\begin{align}
 X_a = \begin{pmatrix}
        {X_a}_{(1)} & 0 & 0 \\
        0 &  {X_a}_{(2)} & 0 \\
        0 &   0 &  {X_a}_{(3)}
       \end{pmatrix}.
\end{align}
This block structure is only manifest in an adapted basis, but it can be detected and reconstructed directly from the configuration \( \{X_a\} \) by analyzing the zero modes of the Laplacian \( \Delta \).

Let \( P_1, \ldots, P_k \) denote the projectors onto the irreducible blocks, satisfying \( P_i P_j = \delta_{ij} P_j \) and \( P_j^\dagger = P_j \). These projectors are zero modes of \( \Delta \), i.e., \( \Delta P_j = 0 \). Conversely, any zero mode \( E_j \) must be a linear combination \( E_j = \sum_k e_{jk} P_k \) of these projectors. Thus, the projectors \( P_k \) can be extracted from the eigenspaces corresponding to zero (or near-zero) eigenvalues.

In practice, the blocks may not be perfectly separated---for instance, due to noise---and one typically encounters several ``almost-zero'' modes \( E_j \) that approximately correspond to irreducible components. Care must be taken in distinguishing these modes from the rest of the spectrum. A complete set of projectors is obtained when the relation \( \sum_j P_j = \one \) is satisfied.

\subsection{Avoiding the curse of dimensionality with high-dimensional quantum geometry}
\label{sec:curse}

High-dimensional data often suffer from the ``curse of dimensionality''---an exponential growth in volume that renders many classical methods inefficient \cite{Bellman1957}. However, most real-world datasets exhibit \emph{concentration of measure} \cite{donoho2000high}, meaning that the data are effectively supported on lower-dimensional manifolds. This motivates dimensionality reduction as a necessity rather than a convenience.

Classical methods such as PCA, Laplacian eigenmaps, and UMAP \cite{Pearson1901, belkin2003laplacian, mcinnes2018umap} aim to recover the intrinsic manifold by analyzing pairwise distances. However, these approaches rely solely on pointwise information, limiting their ability to capture global geometric or topological structure.

Quantum geometry offers a powerful alternative. Here, geometry is encoded not in distances between data points, but in a continuous manifold \( \mathcal{M} \subset \mathbb{C}P^{N-1} \) defined by quasi-coherent states derived from a learned matrix configuration \( \{X_a\} \). The geometry is governed by operator spectra and quasi-classical quantities such as the quantum metric and symplectic form \cite{steinacker2024quantum}.

This representation is remarkably efficient: for example, the minimal fuzzy \( \mathbb{C}P^{N-1} \) discussed in Appendix \ref{app:QG-basic-examples}
encodes a smooth \( 2(N-1) \)-dimensional manifold using only \( N^2 - 1 \) matrices of size \( N \). In general, a compact manifold of intrinsic dimension \( d \) can be captured by fuzzy geometry using only \( d + 1 \) matrices.  At fixed matrix size \(N\), the number of operator parameters grows linearly with the ambient dimension \(D\).

While the minimal configurations are rigid, increasing expressivity through higher harmonics (polynomials in \( X_a \)) is possible with a moderate increase in matrix size. Importantly, increasing the number of features does not require larger matrices unless the new features carry independent geometric content.

More generally, in the extended version of quantum geometry used in QCML, we retain not only Hermitian matrices but also quasi-coherent states corresponding to data points. In this setting, the total number of real parameters describing the representation is \( P = DN^2 + 2T(N - 1) \), where \( D \), \( T \), and \( N \) denote the number of features, the number of data points, and the dimension of the Hilbert space---a hyperparameter of the model. The QCML representation of the data is lossy for small \( N \), when the number of learned parameters is (much) smaller than the full dimensionality of the dataset, i.e., \( P < TD \). Increasing \( N \) reduces the degree of lossiness, with the representation becoming lossless at sufficiently large \( N \) (definitely when \( N \geq T \)). In applications\cite{candelori2025robust,di2025quantum,samson2024quantum,rosaler2025supervised} where important features of the data can be captured with modest values of \( N \), this approach overcomes the curse of dimensionality. 
Remarkably, in practical applications~\cite{candelori2025robust,di2025quantum,samson2024quantum,rosaler2025supervised}, 
values of \( N \) between \( 8 \) and \( 32 \) have proven sufficient to capture the essential geometric and statistical structure of complex datasets. 
QCML thereby replaces vast collections of high-dimensional features with a compact set of operator degrees of freedom, 
 providing a compact representation of the relevant structure. 
In this way, QCML overcomes the curse of dimensionality, with the emergent quantum geometry capturing global data organization through a small number of learned matrices and states.

In summary, quantum geometry—when paired with the concentration of measure—offers a scalable and expressive approach to modeling high-dimensional data. Using this modeling, QCML can efficiently capture curvature, topology, intrinsic dimension, and other geometric characteristics of the data \cite{candelori2025robust}.

\section{Examples}
\label{sec:examples}

The basic example shown in Figure~\ref{fig:1-sphere-N4-E20000} considers data points uniformly distributed on the surface of the unit sphere, a setting with high symmetry. In such cases, it is not surprising that the quantum geometric representation performs exceptionally well.

{ The basic QCML construction and the use of the quantum metric for intrinsic-dimension estimation were introduced in our previous work.\cite{musaelian2024qcml,candelori2025robust} All other analyses and results presented in this section are new, including the extraction of Berry curvature and monopole charges, the study of the matrix-Laplacian spectrum and eigenmaps, and the application of these tools to the datasets considered below.}

In this section, we turn to more complex datasets to demonstrate that QCML continues to produce meaningful quantum geometric representations even in the absence of strong symmetry. We show that key geometric and topological features of the data can be naturally and robustly extracted from the resulting quantum geometry. A summary table of the examples is provided in Appendix~\ref{app:examples-summary}.

\subsection{Two disconnected spheres with noise}
 \label{sec:ex-2-spheres}

Consider two 2D spheres with centers at $x=y=0$ and $z=0,3$ and radii $1.5,1$, respectively. The data will consist of $n=2000$ points generated uniformly over surfaces of these spheres with the standard Gaussian noise of the strength $\eta=0.1$ added to each coordinate of the generated point.

\begin{figure}[htbp]
\centering
\includegraphics[width=\textwidth,height=229.56686000pt]{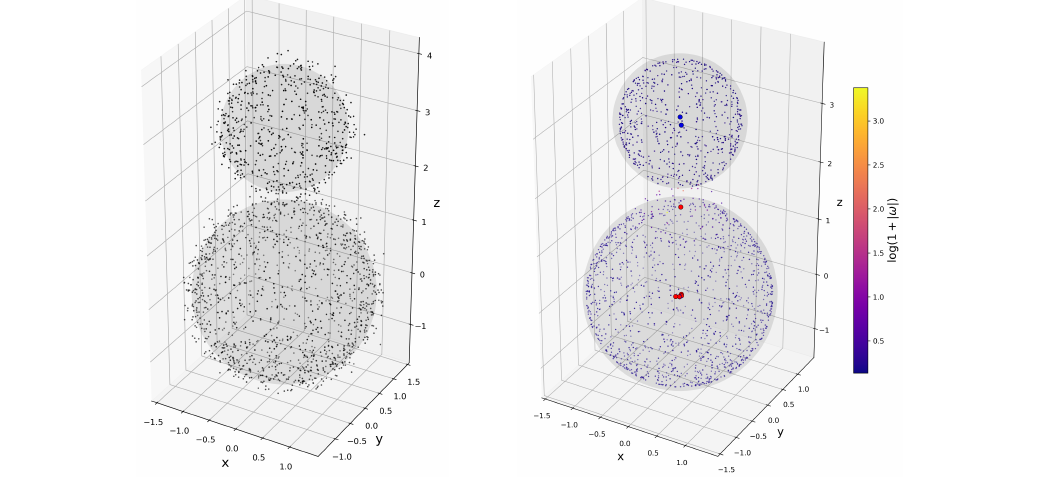}
\caption{Left: Dataset of 2000 points sampled uniformly from the surfaces of two spheres centered at \( z = 0 \) and \( z = 3 \) with radii 1.5 and 1, respectively. The surfaces are shown in gray. Gaussian noise with standard deviation \( \eta = 0.1 \) was added independently to each coordinate of each point. 
    Right: Quantum geometric representation learned via trained quantum operators \( \mathbf{X} \). The QCML point cloud is shown, with point colors indicating the local Berry curvature. Seven degeneracy points of the displacement Hamiltonian (monopoles) in the illustrated run are highlighted in red and blue, corresponding to topological charges \( \pm 1 \), respectively.
}
\label{fig:2spheres-QCML}
\end{figure}
Figure~\ref{fig:2spheres-QCML} shows the original dataset alongside the corresponding QCML cloud, derived from the expectation values of the quantum geometric operators learned during training. The QCML representation is visibly less noisy, with points clustering tightly around the underlying surfaces defined by the learned quantum geometry. In the illustrated run, seven degeneracy points of the displacement Hamiltonian are identified and color-coded by their topological charges. Four monopoles, i.e., degeneracy points with topological charges \( c_1 \neq 0 \), are concentrated near the center of the larger sphere and two near the smaller one—roughly consistent with their relative surface areas. Accordingly, the Berry curvature, indicated by the point colors, is approximately constant across each sphere, as expected. The one notable exception is the region near the seventh monopole, located where the two spheres approach each other. The seventh monopole lies near the neck between the spheres in this particular learned geometry. 

\begin{figure}[htbp]
\centering
\includegraphics[width=\textwidth,height=234.50404000pt]{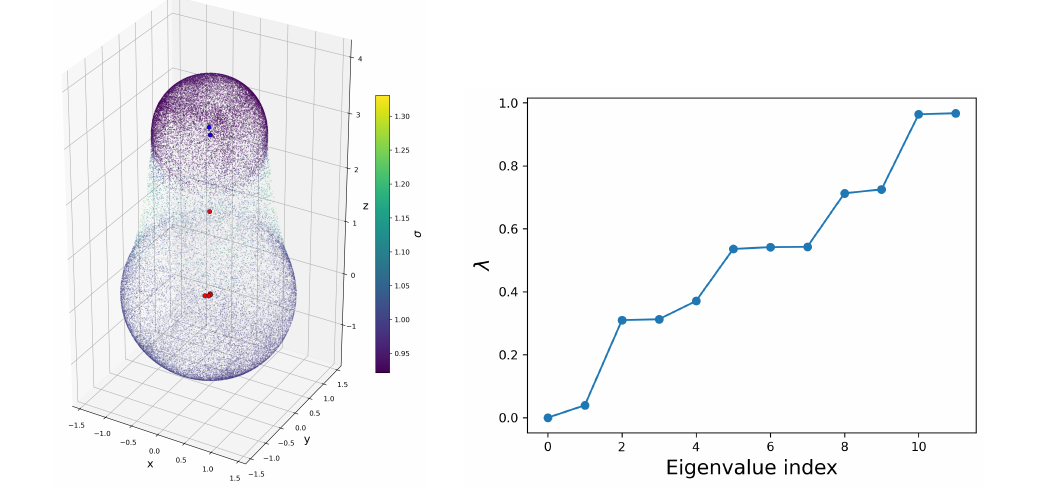}
\caption{Left: Quantum geometry point cloud derived from the learned quantum geometry. Points are colored by the uncertainty \( \sigma(x) \) evaluated at each location. Seven degeneracy points (monopoles) of the displacement Hamiltonian in the illustrated run are highlighted in red and blue, corresponding to topological charges \( \pm 1 \), respectively. Right: Partial spectrum of the matrix Laplacian derived from the learned quantum geometry. The first 12 eigenvalues are shown. A spectral gap separates the first two eigenvalues from the rest, consistent with the presence of two almost disconnected components in the data.}
\label{fig:2spheres-Cloud-uncertainty}
\end{figure}

The QCML cloud encodes both the learned quantum geometry and the original data distribution. To isolate the quantum geometry itself, we visualize the corresponding point cloud in Figure~\ref{fig:2spheres-Cloud-uncertainty}.

The illustrated quantum geometry consists of two nearly perfect spheres connected by a bridge. The uncertainty is visibly higher in the bridge region. This point cloud can be viewed as a quantum geometric model of the data.

Figure~\ref{fig:2spheres-Cloud-uncertainty} shows the spectrum of the matrix Laplacian constructed from the learned quantum geometry. The eigenvalue structure reveals important topological and geometric features of the data manifold. The lowest two eigenvalues are nearly degenerate, indicating that the data consist of two almost disconnected components. Moreover, the next six eigenvalues appear in triples, corresponding to the triply degenerate $l=1$ modes on the first and second spheres.

We also used this example to test the dependence on random initialization and on the parameter $w$. The broad two-sphere geometry persists across the tested seeds, but at $w=0.1$ the detected monopole count varies from 3 to 8 across training and dataset seeds, with varying charges. At $w=0.3$, the geometry and low-lying Laplacian spectrum are more consistent across training seeds, although the monopole count still varies from 5 to 7. The corresponding comparison scripts and trained operators are included in the public repository\cite{hogan2026quantumgeometrydata}.

Simultaneous complex conjugation of the learned matrices leaves the loss and expectation-value cloud unchanged but reverses the Berry curvature and all monopole charges. Their overall sign is therefore not fixed by the classical data.

\subsection{A sphere with non-uniform distribution of points}
 \label{sec:ex-nonuniform-sphere}

As our next example, consider a dataset consisting of 2000 points distributed on the surface of a unit two-dimensional sphere, with a probability density proportional to \( (1 + \cos\theta)^3 \) with respect to the surface-area measure. This distribution is denser near the north pole of the sphere. We also add standard Gaussian noise with strength \( \eta = 0.1 \) to each coordinate of the generated points.

\begin{figure}[htbp]
\centering
\includegraphics[width=\textwidth,height=190.86916000pt]{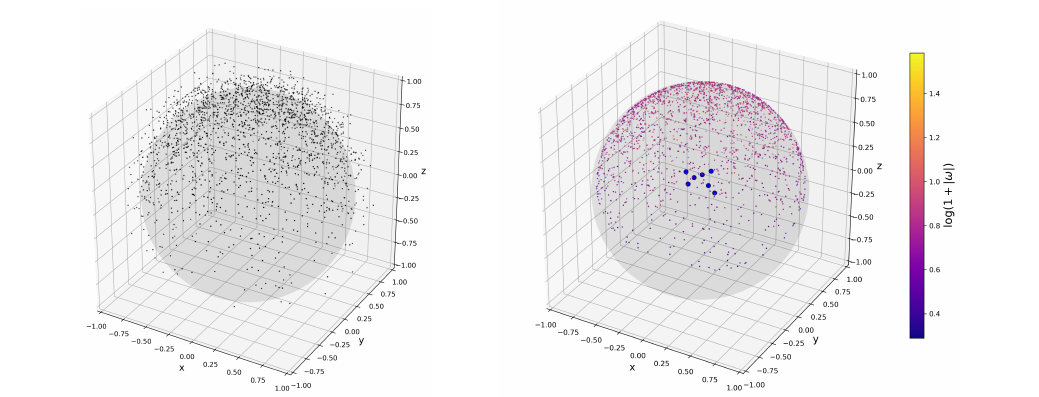}
\caption{Left: Dataset of 2000 points sampled on the surface of a 2D unit sphere with probability density proportional to $(1+\cos\theta)^3$ with respect to surface area. The unit sphere is shown in gray for reference. Gaussian noise with standard deviation \( \eta = 0.1 \) was added independently to each coordinate of each point. 
    Right: Quantum geometric representation learned via trained quantum operators \( \mathbf{X} \). The QCML point cloud is shown, with point colors indicating the local Berry curvature. Seven degeneracy points of the displacement Hamiltonian (monopoles) are highlighted in blue, corresponding to topological charges \( - 1 \).
}
\label{fig:1spheres-nonuniform-QCML}
\end{figure}
\begin{figure}[htbp]
\centering
\includegraphics[width=\textwidth,height=174.59132000pt]{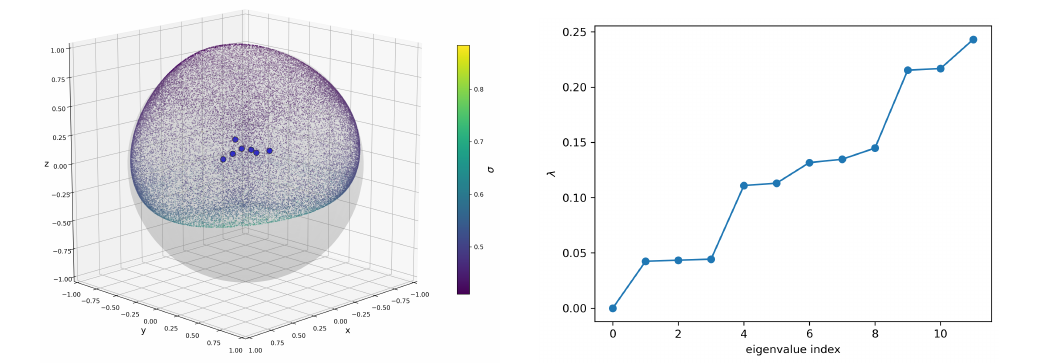}
\caption{Left: Quantum geometry point cloud derived from the learned quantum geometry. Points are colored by the uncertainty \( \sigma(x) \) evaluated at each location. Seven degeneracy points (monopoles) of the displacement Hamiltonian are highlighted in blue, corresponding to topological charge \( -1 \). Right: Partial spectrum of the matrix Laplacian derived from the learned quantum geometry. The first 12 eigenvalues are shown. The clustering of eigenvalues into groups of 1, 3, and 5 reflects the characteristic spectrum of the Laplacian on a sphere, corresponding to angular momentum modes \( \ell = 0, 1, 2 \), respectively.}
\label{fig:1spheres-nonuniform-Cloud-uncertainty}
\end{figure}

Figure~\ref{fig:1spheres-nonuniform-QCML} shows the original dataset alongside the corresponding QCML cloud, derived from the expectation values of quantum geometric operators learned during training. 
To visualize the learned quantum geometry in isolation from the original data, we present the corresponding point cloud in Figure~\ref{fig:1spheres-nonuniform-Cloud-uncertainty}.

The resulting quantum geometry forms a deformed spherical surface. The upper region closely resembles a unit sphere, while the lower region is visibly compressed. Notably, the uncertainty is higher in the lower part of the surface, where the original data was sparsely sampled. This point cloud serves as a quantum geometric model of the data: the learned geometry fills in missing regions with coherent cloud points while preserving elevated quantum uncertainty in areas of low data density.

We also show in 
Figure~\ref{fig:1spheres-nonuniform-Cloud-uncertainty} the spectrum of the matrix Laplacian constructed from the learned quantum geometry. The clustering of eigenvalues into groups of 1, 3, and 5 reflects the characteristic spectrum of the Laplacian on a sphere, corresponding to angular momentum modes \( \ell = 0, 1, 2 \), respectively. We also checked that the Weyl approach to investigate the intrinsic dimension outlined in Section~\ref{sec:matrix-Laplacian}, is consistent with an intrinsic dimension of the data manifold $d = 2$.

\subsection{Dataset of conformal maps}
 \label{sec:ex-conformal-maps}

The next synthetic example demonstrates how QCML handles high-dimensional data with low intrinsic dimension. We first consider a dataset consisting of conformal maps from the unit disk in \( \mathbb{C} \) to itself, given by
\begin{align}
    z \mapsto f(z\,|\,a,\theta) = e^{i\theta}\frac{a - z}{1 - \bar{a} z},
 \label{mobius}
\end{align}
where each map is parametrized by an angle \( \theta \in [0, 2\pi) \) and a complex number \( a \) that satisfies \( |a| < 1 \). After examining conformal self-maps of the unit disk, we will generalize to datasets consisting of analytic self-maps of the unit ball in \( \mathbb{C}^n \) for n=2, 3, 4, and 5. While the space of all such maps is infinite-dimensional, we construct a finite dataset by discretizing these maps as follows.

We begin by sampling 100 fixed reference points \( \{(\tilde{x}_i, \tilde{y}_i)\} \) uniformly from the interior of the unit disk. For given parameters \( a \) and \( \theta \), we compute the image of these reference points under the map \( f(z\,|\,a,\theta) \), yielding a set of transformed points \( \{(x_i, y_i)\} \) with \( x_i + i y_i = f(\tilde{x}_i + i \tilde{y}_i\,|\,a,\theta) \). 
We consider the point $(x_1, y_1, x_2, y_2, \ldots, x_{100}, y_{100})$ as a data point in $\mathbb{R}^D$ with $D=200$ that approximately encodes the conformal map $f(z|a,\theta)$. 

For simplicity, we fix \( \theta = 0 \) (i.e., no global rotation) and sample \( a \) uniformly from the disk of radius \( r_{\text{max}} = 0.9 \). This yields a set \( \{a^t\} \) for \( t = 1, \ldots, T = 2000 \), from which the corresponding data points are computed. The resulting dataset consists of \( 2000 \) conformal maps \( \{f^t\}_{t=1}^T \), each represented by 200 real coordinates \( x_i^t, y_i^t \). The full dataset forms a \( T \times D \) matrix with \( D = 200 \) and \( T = 2000 \), which serves as the input to QCML.

The advantage of this set is that we know that the data points concentrate near two-dimensional surface parametrized by $a$ and embedded into 200-dimensional space. 

\begin{figure}[htbp]
\centering
\includegraphics[width=\textwidth,height=125.26376000pt]{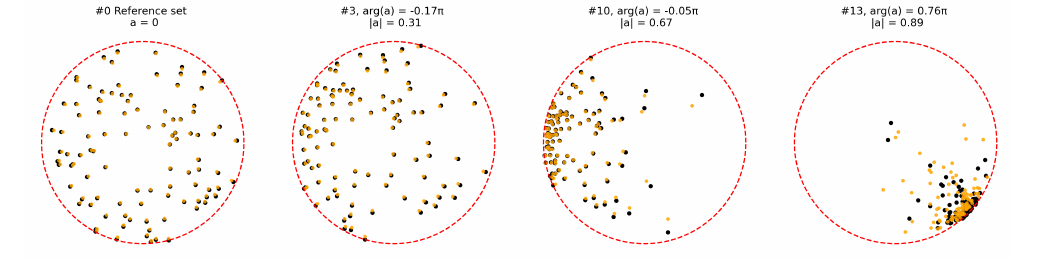}
\caption{Examples of conformal maps from the dataset. Each map is represented by the image (black points) of 100 reference points (shown in the leftmost panel) under a conformal transformation \( f(z; a^t, \theta = 0) \), where \( a^t \) is sampled uniformly from the disk \( |a| < 0.9 \). The corresponding QCML-learned quantum geometric representations are shown in orange. The associated complex parameters \( a \) are displayed in polar coordinates.}
\label{fig:conformal-map-examples}
\end{figure}

The Figure~\ref{fig:conformal-map-examples} shows examples of data points. Each data point is shown as 100 black points inside the unit disk. The points have been obtained as the image of 100 reference points under conformal transformation (\ref{mobius}) with $\theta=0$ and randomly generated $a$.  We performed QCML training to the dataset with $N=8, w=0.1$. The corresponding QCML cloud is shown as orange points in the same figure. The QCML points are pretty close to the original points confirming that learning has been successful. 

\begin{figure}[htbp]
\centering
\includegraphics[width=\textwidth,height=152.80116000pt]{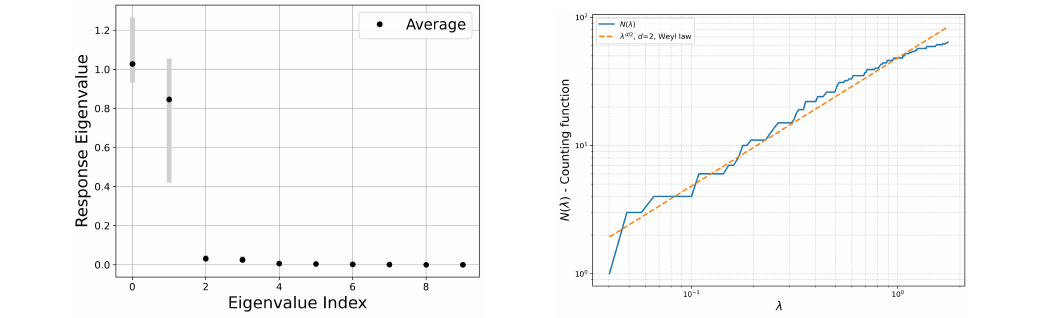}
\caption{
    Left: Spectrum of the response matrix at all data points for the conformal map dataset. Gray lines show the spread of eigenvalues; black dots indicate the mean. The gap after the first two eigenvalues suggests intrinsic dimension \( d = 2 \). 
    Right: Spectral counting function \( N(\lambda) \) vs eigenvalue \( \lambda \) of the matrix Laplacian. The approximate linear growth of \( N(\lambda) \sim \lambda^{d/2} \) is consistent with  \( d = 2 \) as an intrinsic dimension of the data.
    }
\label{fig:conformal-metric-spectral}
\end{figure}

To extract the intrinsic dimension of the dataset from QCML we use the approach of Ref.~\citeonline{candelori2025robust} and plot the spectrum of the response matrix (see Section~\ref{sec:QGstructure}) evaluated at every point of QCML cloud. The plot is shown in the left panel of Figure~\ref{fig:conformal-metric-spectral} and demonstrates clear gap between the top two eigenvalues and the rest signifying the intrinsic dimension $d=2$ of the data. 
The matrix Laplacian spectrum is shown in the right panel of Figure~\ref{fig:conformal-metric-spectral} as a spectral counting function $N(\lambda)$. The latter is defined as the number of eigenvalues lower than $\lambda$. This is a log-log plot illustrating that the spectrum is consistent with the Weyl's law with $d=2$ (see Section~\ref{sec:matrix-Laplacian}). { Because of the finite matrix cutoff, the gap in the response spectrum
is used as the primary dimension diagnostic and the Weyl-law fit as a consistency
check.}

\begin{figure}[htbp]
\centering
\includegraphics[width=\textwidth,height=198.29047000pt]{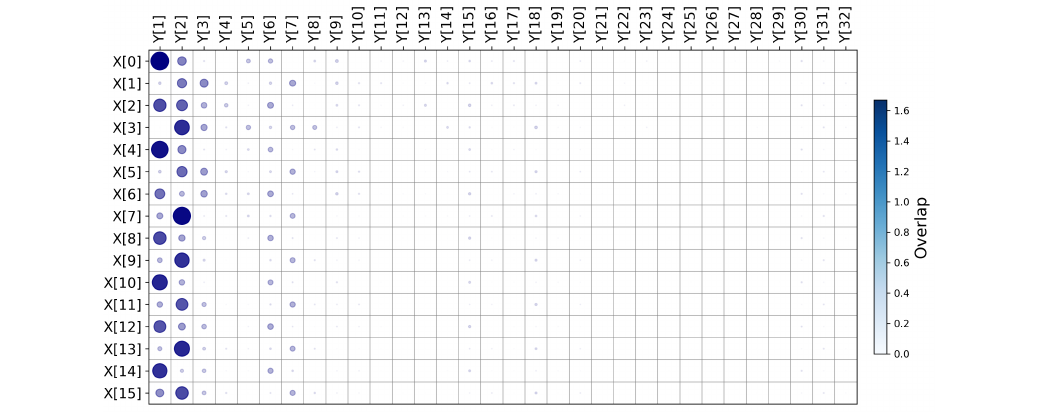}
\caption{Overlap matrix between the learned quantum observables \( X_a \) and the eigenmodes \( Y_i \) of the matrix Laplacian. The size and intensity of circles indicate the magnitude of the overlap \( \operatorname{Tr}(Y_i X_a) \). Only a subset of the eigenmodes is shown. The strong alignment of  \( X_a \) with only first two leading eigenmodes \( Y_{1,2} \) suggests that these correspond to the dominant geometric directions in the feature space. Only a portion of the full overlap matrix is shown.}
\label{fig:XY-overlap}
\end{figure}

Another important tool of QCML data analysis is matrix Laplacian eigenmaps (Section~\ref{sec:eigenmaps}). We compute the eigenmodes $Y_k$ of the matrix Laplacian. The lowest mode $Y_0$ is always proportional to a unity matrix. We find the overlap coefficients $\tr(Y_k X_i)$ between the Hermitian matrices $Y_k$ and the feature matrices $X_i$ learned during QCML training. The obtained overlap matrix is presented as a heatmap in Figure~\ref{fig:XY-overlap}. The size and intensity of circles indicate the magnitude of the overlap. It is clear from the figure that the first two modes $Y_{1,2}$ clearly dominate. It is not surprising that they correlate with the $x$ and $y$ features of data points as can be seen from the chessboard pattern of overlaps.   

\begin{figure}[htbp]
\centering
\includegraphics[width=\textwidth,height=169.59221000pt]{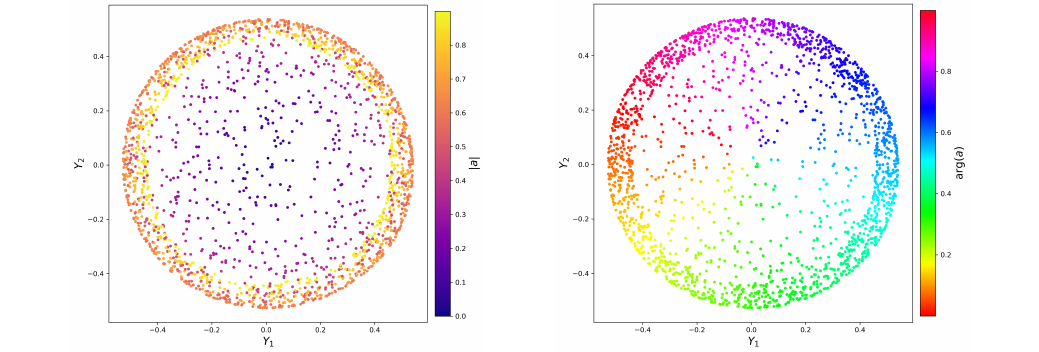}
\caption{QCML cloud for the conformal map dataset visualized using the first two Laplacian eigenmaps \( (Y_1, Y_2) \). Left: Points are colored by the modulus \( |a| \) of the complex parameter generating the map.  Right: Points are colored by the phase \( \arg(a) \).}
\label{fig:conformal-eigenmap-colored}
\end{figure}

To have even better understanding on how QCML ``learned'' geometry of the data we plot the QCML cloud projected onto $Y_1-Y_2$ matrix geometry. The Figure~\ref{fig:conformal-eigenmap-colored} shows a scatter plot of points $\bra{x^t}Y_{1,2}\ket{x^t}$ computed using quasi-coherent states $\ket{x^t}$ for dataset points. We color the points by $|a^t|$ (left panel) and by $\arg(a^t)$ (right panel), where $a^t$ is a complex parameter of a conformal map corresponding to the point. While the QCML training was not supervised so that values $a^t$ were not used, there is a clear color pattern in Figure~\ref{fig:conformal-eigenmap-colored}. One can roughly think about $Y_{1,2}$ corresponding to real and imaginary parts of $a$ so that $Y_1+i Y_2 \sim a$. 

\begin{figure}[htbp]
\centering
\includegraphics[width=\textwidth,height=180.25429000pt]{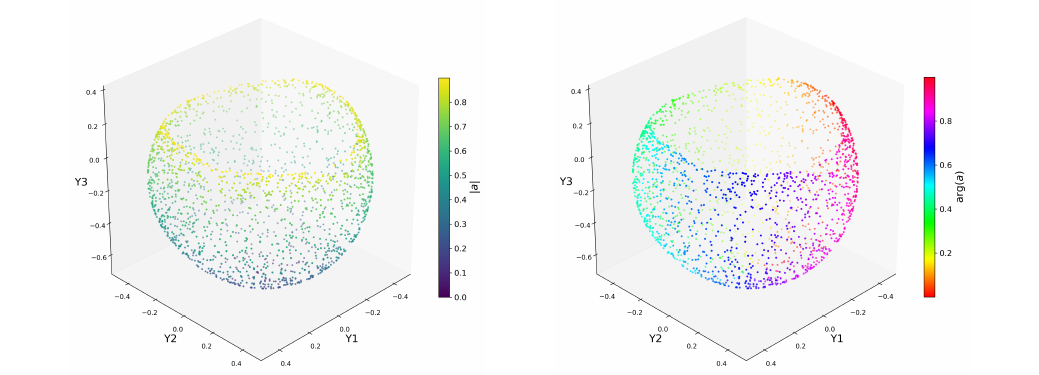}
\caption{Quantum geometry of the conformal map dataset visualized via Laplacian eigenmaps \( (Y_1, Y_2, Y_3) \). Left: Points colored by modulus \( |a| \) of the conformal parameter. Right: Same embedding, colored by argument \( \arg(a) \).}
\label{fig:conformal-3D-colored}
\end{figure}

As it can be seen from the matrix overlap in Figure~\ref{fig:XY-overlap}, the component $Y_3$ is also significant. We therefore plot the 3D projection of the QCML cloud in Figure~\ref{fig:conformal-3D-colored}. It shows a 2D surface embedded in the 3D space. The surface has a topology of a disk and preserves the rotational symmetry of the disk.  Figure~\ref{fig:conformal-eigenmap-colored} is a 2D projection of the 3D one and loses some important structure (e.g., notice the folding of the surface under the projection close to the disk boundary). The original dataset contains not only the geometry of $a$-disk but also important information specific to the embedding into high-dimensional feature space. This information is encoded into higher eigenmaps. The eigenmaps with very high eigenvalues roughly correspond to higher harmonics of the data distribution and are sensitive to the noise caused by statistical nature of this synthetic dataset.

The conformal maps of the unit disk $\mathbb{B}^1$ considered here, can be generalized to the analytic self-maps of a unit ball \(\mathbb{B}^n  \subset  \mathbb{C}^n \), for arbitrary integer $n>1$.  These are known as the Blaschke-Potapov factors \cite{potapov1960multiplicative},
\begin{align}
    z \mapsto f(z\,|\,a,\theta) = e^{i\theta}\sqrt{1-\|a\|^2}\frac{a - z}{1 - \langle a|z \rangle}(I_n-a a^\dagger )^{-\frac12}.
 \label{BPfactors}
\end{align}
Here \(z,a\in\mathbb{B}^n\), \(\theta\in[0,2\pi)\), \(\|a\|\) represents the norm of \(a\), { and
$
\langle a|z\rangle=a^\dagger z
$
is the standard Hermitian inner product. The factor
$
(I_n-aa^\dagger)^{-1/2}
$
denotes the inverse square root of the positive-definite matrix
\(I_n-aa^\dagger\). For \(n=1\), one has
\(\langle a|z\rangle=\bar a z\), and Eq.~\eqref{BPfactors} reduces to the scalar
Möbius transformation in Eq.~\eqref{mobius}.}
We generated datasets  using (\ref{BPfactors}) with random  $a$-s for $n=2, 3, 4$, and $5$. We applied QCML training to these Blaschke-Potapov map datasets. The intrinsic dimension was correctly computed at all sample points for all tested dimensions.

To conclude, this example demonstrates that QCML is effective at learning symmetries, topology and other geometric features of higher-dimensional datasets without manually adding inductive biases catered to conformal maps or other analytic structures. As a result, QCML is a natural tool to analyze datasets that were created by a potentially exotic generation process. Such datasets include model weights from neural network training, complex sensor networks in medical imaging and measurement, and scientific data with many features. 

\subsection{Wisconsin Breast Cancer Data}
\label{sec:ex-WBC}

{ We also evaluate our methodology on the Diagnostic Wisconsin Breast Cancer Database (WBC) \cite{misc_breast_cancer_wisconsin_(diagnostic)_17}. The dataset comprises cases from women presenting with breast lesions that were clinically suspicious for malignancy. All subjects underwent fine-needle aspiration cytology (FNAC), and definitive benign or malignant classifications were established through histopathological examination. Because all participants were referred for biopsy based on abnormal clinical or imaging findings, the benign cohort should not be interpreted as a healthy control population, but rather as individuals with lesions ultimately determined to be noncancerous. This dataset contains 569 data points, each representing an image obtained by FNAC of a breast tumor. From each image, $D=30$ features are extracted to characterize the properties of the cell nuclei present. We have $T = 569$ samples lying on a manifold embedded in a 30-dimensional Euclidean space.  Ten real-valued features are computed for each cell nucleus: radius (mean of distances from center to points on the perimeter);  texture (standard deviation of grayscale values); perimeter; area; smoothness (local variation in radius lengths); compactness ($\text{perimeter}^2 / \text{area} - 1.0$); concavity (severity of concave portions of the contour); concave points (number of concave portions of the contour); symmetry; fractal dimension (``coastline approximation'' - 1).  For each image, the mean, standard error, and ``worst'' value (the mean of the three largest values) were calculated for each of the ten measurements, resulting in 30 features. For example, the mean, standard error, and worst of fractal dimension represent different aspects of the fractal dimension measurements. The 30 features vary widely in scale, with standard deviations ranging from $2.65 \times 10^{-3}$ to $5.69 \times 10^2$. To account for this, we apply standard scaling to normalize each feature to have zero mean and unit variance.  

To apply QCML analysis to the data we have to choose the values of hyperparameters $N$ and $w$. 
Given the ambient dimensionality $D=30$, we choose a Hilbert-space dimension of $N=8$. For the loss function, we incorporate a quantum fluctuation term with weight factor $w$ ranging from $w = 0.02$ to $w=0.8$, as specified in equation~\eqref{eq:qcmlloss}.

For reference, classical principal component analysis (PCA) identifies the directions in which the data vary the most and, when applied to the WBC dataset, typically reveals that over 90\% to 95\% of the variance can be captured in just 7 to 10 principal components. This suggests that while the data exist in a 30-dimensional space, its meaningful structure lies largely within a much lower-dimensional subspace.

For these data, following the QCML approach described in Ref.~\citeonline{candelori2025robust}, 
we compute the average eigenvalues of the response matrix for different values of the parameter~$w$. 
When the learned expectation-value map approximates projection onto a
\(d\)-dimensional manifold, \(d\) response eigenvalues are near unity and the
others are near zero. 
For real or noisy datasets, however, this separation is no longer sharp~\cite{candelori2025robust}. 
As an estimate of the intrinsic dimension, we count the number~$d$ of averaged eigenvalues exceeding~$0.5$. 
The results are summarized in Table~\ref{tab:intrinsic-dimension}.

\begin{table}[h!]
\centering
\caption{Estimated intrinsic dimension $d$ as a function of the parameter $w$. 
The values of $d$ are computed as the number of averaged response eigenvalues greater than the conventional threshold of~$0.5$.}
\label{tab:intrinsic-dimension}
\begin{tabular}{c|cccccc}
\hline
$w$ & 0.02 & 0.05 & 0.10 & 0.20 & 0.50 & 0.80 \\
\hline
$d$ & 7    & 5    & 4    & 3    & 2    & 1    \\
\hline
\end{tabular}
\end{table}

As discussed in Section~\ref{sec:QCML}, there is no universal prescription for choosing the value of~$w$. In fact, analyzing the data for different values of the parameter~$w$, which controls the suppression of quantum fluctuations, can reveal different features of the dataset. Let us begin by focusing on the intermediate value $w=0.2$.

To gain a more detailed insight into the geometry of the lower-dimensional space and its embedding, we compute the spectrum of the matrix Laplacian. The Laplacian spectrum for $w=0.2$, shown in the left panel of Figure~\ref{fig:WBC}, does not exhibit degenerate zero modes, indicating the absence of disconnected components in the data. Moreover, the response spectrum indicates an intrinsic dimension of approximately three. As a consistency check, a Weyl-law fit over one logarithmic decade centered on the scale \(\lambda_{\rm NC}\) defined in Section~\ref{sec:matrix-Laplacian}, with the trivial zero mode excluded, gives \(d\approx2.5\). Given the limited spectrum at \(N=8\), the Weyl-law estimate should be interpreted cautiously.

In the right panel of Figure~\ref{fig:WBC}, we show the overlap matrix \( \operatorname{Tr}(Y_i X_a) \) between the learned observables \( X_a \), corresponding to the 30 WBC features, and the Laplacian eigenmaps \( Y_i \) computed from Eq.~(\ref{eigenmaps-100}). Only the portion of the matrix corresponding to eigenmaps \( Y_1 \) through \( Y_{31} \) is shown; the zero mode \( Y_0 \), which is proportional to the identity matrix, is omitted. Eigenmaps with lower eigenvalues tend to capture more global properties of the data and offer valuable insights into feature relevance. This analysis may be viewed as dual to conventional PCA.

\begin{figure}[htbp]
\centering
\includegraphics[width=\textwidth,height=176.05324000pt]{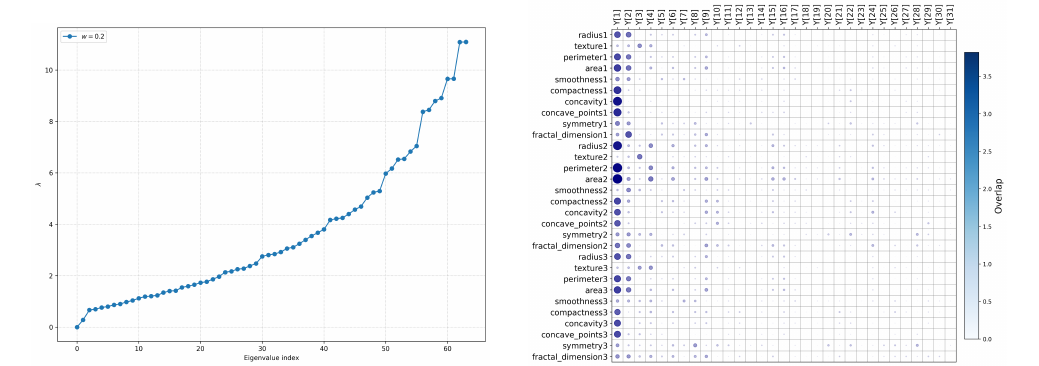}
\caption{{
    Left: Spectrum of the normalized matrix Laplacian $\Delta/30$ derived from the learned quantum
    geometry at $w=0.2$. The absence of degenerate zero modes indicates that
    the data geometry has no disconnected components.
    Right: Overlap matrix between the learned quantum observables $X_a$ (rows),
    corresponding to 30 features of the Wisconsin Breast Cancer Data, and the
    eigenmodes $Y_i$ of the matrix Laplacian (columns). The size and intensity
    of circles indicate the magnitude of the overlap
    $\operatorname{Tr}(Y_i X_a)$. Only a subset of the eigenmodes is shown.
    }}
\label{fig:WBC}
\end{figure}

The learned quantum geometry of the data set can be visualized using the leading Laplacian eigenmaps 
\((Y_1, Y_2, Y_3)\), as shown in Figure~\ref{fig:WBC2}. 
The resulting point cloud is consistent with an intrinsic dimension of approximately \(d \approx 3\). 
The points are colored according to the diagnostic labels provided in the WBC dataset: 
red for malignant cells and green for benign ones. 
Note that these diagnostic labels were not used during the unsupervised QCML training described in this section. 

It is instructive to compare the quantum-geometric description with a purely
classical analysis of the same dataset. This pipeline reaches its main
finding in three stages. 

First, the global PCA on the standardized data matrix fixes a
linear subspace. The eigenvalue spectrum decays smoothly, and the choice of
subspace dimension is criterion-dependent: the Kaiser rule
(keep sample eigenvalues $\geq 1$) gives $d = 6$, the largest consecutive eigenvalue ratio
gives $d = 2$, and cumulative-variance thresholds give $10$ at $95\%$ and $17$
at $99\%$. Adopting the Kaiser value retains $88.8\%$ of the variance in six
components.

Second, varimax rotation within that six-dimensional subspace replaces the
principal axes, each of which mixes many original features, by six factors of
simple structure, each loading heavily on a small subset of features. The
factors are biologically legible: tumor size ($F_1$, from area, radius, and
perimeter); boundary irregularity ($F_2$, from the standard-error variants of
compactness, concavity, concave points, and fractal dimension); measurement
scatter ($F_3$, from the standard-error variants of smoothness, texture, and
symmetry); gray-level texture ($F_4$); surface smoothness and fractal complexity
($F_5$, from the mean and worst variants of smoothness and fractal dimension);
and nuclear symmetry ($F_6$). Two conventions should be noted: the loadings on
$F_1$ and $F_2$ are negative, so that high $F_1$ corresponds to a smaller tumor
and high $F_2$ to a more regular boundary; and the standard-error variants of a
base measurement generally load on different factors than the mean and worst
variants, which is why fractal dimension appears in both $F_2$ and $F_5$.
Rotation preserves the subspace and the variance that it explains; it
redistributes them into coordinates that allow interpretation.

Third, the local PCA within the rotated space, fitting a sample covariance
to the $K = 15$ nearest neighbors of each point and locating the largest
eigenvalue ratio, returns a modal local dimension $d = 5$. Decomposing the mean local normal direction over the flattest
decile of points attributes $50.8\%$, $22.6\%$ and $19.4\%$ of its variance to
$F_3$, $F_2$ and $F_5$ respectively, so that these three roughness factors
account for about $93\%$ of the locally compressed direction, while texture and
symmetry contribute negligibly.  These values are not directly comparable with the estimate above. The local
dimension is measured within a linear embedding at a fixed neighborhood size,
whereas the count from the response spectra is global and depends on the resolution
parameter $w$, returning $d = 5$ at $w = 0.05$ and $d = 3$ at $w = 0.2$
(Table~\ref{tab:intrinsic-dimension}). We emphasize that the operator framework does not
by itself yield a single preferred dimension for these data: the values obtained
across $w \in [0.02, 0.8]$ span much the same interval as the classical criteria
listed above. What it supplies is a resolution parameter with a geometric
meaning, in place of a choice among rules with no common scale.  The quantum-geometric description requires no such staging: nonlinearity is present from the outset, and intrinsic dimension, connectivity, and feature scale structure follow as spectral readouts of a single learned configuration.

Figure~\ref{fig:WBC2} shows visible separation between benign and malignant samples in
the $(Y_1,Y_2,Y_3)$ coordinates, obtained without label supervision. To
quantify this, we evaluated two classifier families: $\ell_2$-regularized
logistic regression and gradient boosting using ten-fold stratified
cross-validation, comparing the eigenmap coordinates with matched numbers of
principal components and against the entire set
of characteristics (Table~\ref{tab:classifiers}).
Sensitivity and specificity are computed at the default decision threshold of \(0.5\) on the predicted probability. The quantities in Table~\ref{tab:classifiers} are reported as means and standard deviations across folds.

\begin{table}[h]
  \centering
  \begin{tabular}{llcc}
    \hline
    Representation & Classifier & ROC AUC & Accuracy \\
    \hline
    Eigenmaps, $w=0.2$ (3 modes) & Logistic regression & {$0.973 \pm 0.019$} & {$0.907 \pm 0.046$} \\
    Eigenmaps, $w=0.2$ (3 modes) & Gradient boosting   & {$0.981 \pm 0.014$} & {$0.938 \pm 0.024$} \\
    Eigenmaps, $w=0.5$ (3 modes) & Logistic regression & {$0.964 \pm 0.022$} & {$0.912 \pm 0.038$} \\
    Eigenmaps, $w=0.5$ (3 modes) & Gradient boosting   & {$0.970 \pm 0.026$} & {$0.930 \pm 0.032$} \\
    PCA (3 components)           & Logistic regression & $0.990 \pm 0.010$ & $0.948 \pm 0.026$ \\
    PCA (3 components)           & Gradient boosting   & $0.985 \pm 0.015$ & $0.942 \pm 0.025$ \\
    PCA (6 components)           & Gradient boosting   & $0.989 \pm 0.012$ & $0.956 \pm 0.026$ \\
    All 30 features              & Gradient boosting   & $0.992 \pm 0.009$ & $0.962 \pm 0.023$ \\
    \hline
  \end{tabular}
  \caption{{Cross-validated discrimination of benign from malignant samples. Values are means $\pm$ one standard deviation across ten stratified folds.}}
  \label{tab:classifiers}
\end{table}

Three observations follow. First, three unsupervised eigenmap coordinates
reach $0.981$ AUC at $w=0.2$, retaining strong class discrimination after
compression from thirty features. Second, at a matched dimension, the mean
scores remain below those of three principal components, although the gap is
small for gradient boosting: accuracies of $0.938$ versus $0.942$ at $w=0.2$.
Third, gradient boosting increases the mean accuracy for the eigenmaps from
$0.907$ to $0.938$ at $w=0.2$ and from $0.912$ to $0.930$ at $w=0.5$,
suggesting that nonlinear classification can recover additional discriminative
information.

We regard this outcome as expected rather than anomalous. PCA and the
matrix-Laplacian eigenmaps organize the data according to different geometric
criteria: PCA orders linear directions by global variance, whereas the
eigenmaps are ordered by smoothness on the learned matrix geometry, with the
low-lying modes emphasizing larger-scale structure. The parameter $w$ further
controls the resolution at which this geometry is learned. Because the
eigenmaps are selected without reference to the diagnostic labels, there is no
reason for them to maximize class discrimination. We also note that three principal
components already essentially match the full feature set ($0.990$ against
$0.992$ AUC), so the diagnostic signal in this dataset is low-dimensional
and close to linear, leaving little scope for any alternative
representation to demonstrate an advantage on this task. The present example
accordingly supports claims about geometric representation and feature
structure, and does not support a claim of diagnostic advantage; a dataset with
nontrivial topology, or with classes that are not linearly separable, would be
required to test the latter.

\begin{figure}[htbp]
\centering
\includegraphics[width=\textwidth,height=230.22244000pt]{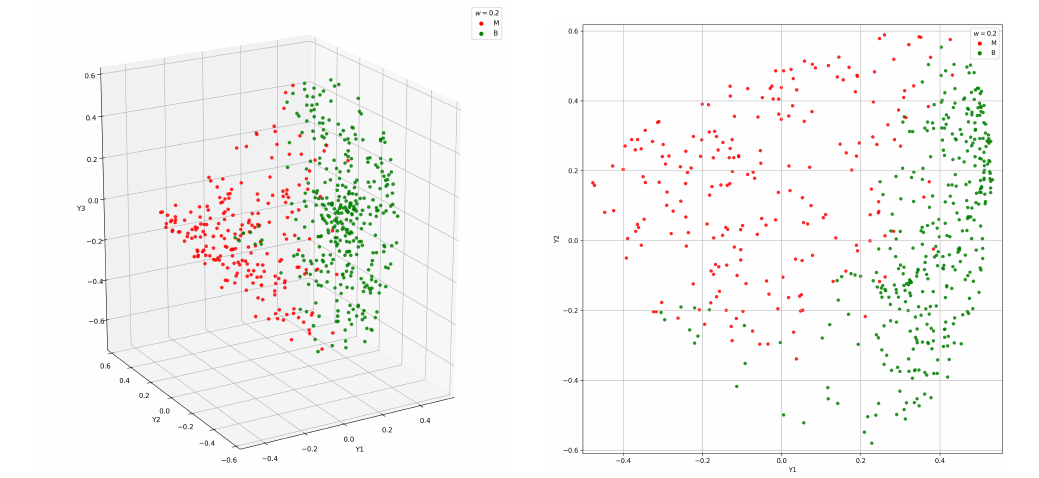}
\caption{{
    Quantum geometry of the WBC dataset at $w=0.2$, visualized using Laplacian
    eigenmaps $(Y_1,Y_2,Y_3)$.
    Left: points colored by diagnosis labels (red for malignant and green for
    benign samples).
    Right: projection onto the two leading eigenmaps $(Y_1,Y_2)$.
    }}
\label{fig:WBC2}
\end{figure}

For comparison, we present a similar eigenmap visualization for $w = 0.5$ in the left panel of Figure~\ref{fig:WBC3}. 
We observe that the expectation values of the lowest eigenmap, $Y_1$, collapse to only two distinct values. 
All samples except two nearly coincident malignant samples lie on a nearly two-dimensional plane $Y_1\approx 0.134$ in the \((Y_1, Y_2, Y_3)\) space. The two samples are separated by only $3.67\times10^{-7}$ in the three-dimensional eigenmap and therefore appear as one visible isolated point.
This observation is consistent with the overall estimate of the intrinsic dimension, \(d \approx 2\), for this value of~$w$. 
The right panel of Figure~\ref{fig:WBC3} shows the spectrum of the matrix Laplacian. 
The first three eigenvalues are very small, indicating that the learned quantum geometry consists of three nearly disconnected components.
The features distinguishing benign from malignant tumors are not arbitrary; they mirror the same morphological abnormalities, such as nuclear enlargement, irregular borders, and variability in cell size or texture, that prompt clinical concern. These traits, more pronounced in malignant cases, reflect the fundamental hallmarks of cancer: nuclear atypia, structural irregularity, and loss of cellular organization. In a separate analysis of the malignant and benign groups (not shown), the first eigenmode dominates in both, but the second eigenmode is appreciable for the malignant group and close to negligible for the benign one. 

An important caution applies to this interpretation. The near-zero modes at
$w = 0.5$ occur where Table \ref{tab:intrinsic-dimension} gives $d \approx 2$, that is, deep in the coarse-grained regime in which a large fluctuation weight suppresses quantum
fluctuations and drives the configuration towards commutativity. As noted in
Section \ref{sec:QCML}, the commutative limit degenerates into $N$-means clustering, whose
matrix Laplacian possesses $N$ exact zero modes, and in which the quantum
geometry is lost entirely. The three low eigenvalues may therefore reflect
genuine reducibility of the data geometry or an early symptom of this collapse.  We regard the three-component reading at $w = 0.5$ as provisional.

At $w=0.5$, logistic regression of the eigenmap coordinates attains
mean specificity $0.952$ and mean sensitivity $0.845$ across the ten stratified
folds.

\begin{figure}[htbp]
\centering
\includegraphics[width=\textwidth,height=200.18915000pt]{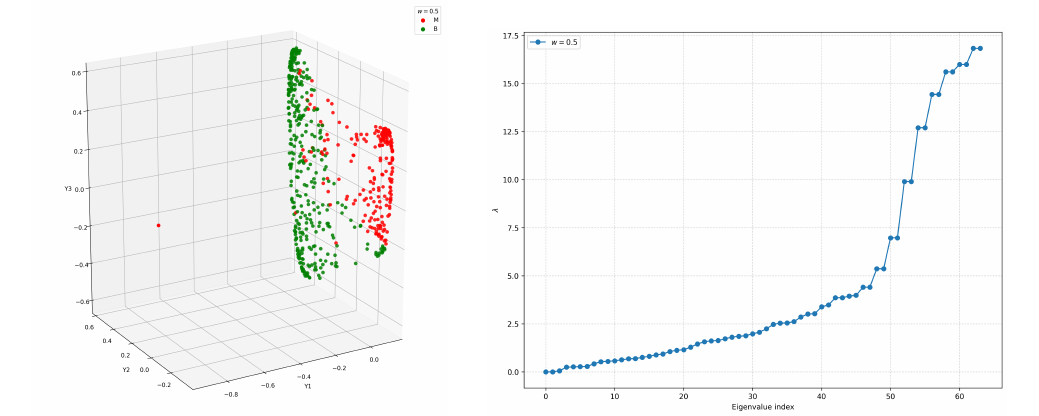}
\caption{{
    Left: Quantum geometry of the WBC dataset visualized using Laplacian eigenmaps
    $(Y_1,Y_2,Y_3)$.
    The expectation values of $Y_1$ collapse onto two distinct levels, placing
    567 samples on a nearly two-dimensional surface and two nearly coincident
    malignant samples at one visible isolated location.
    Points are colored by diagnosis labels (red for malignant and green for
    benign samples).
    Right: Spectrum of the normalized matrix Laplacian $\Delta/30$ derived from the learned quantum
    geometry. The lowest part of the spectrum contains three very small
    eigenvalues, indicating that the quantum geometry consists of three nearly
    disconnected components.
    }}
\label{fig:WBC3}
\end{figure}

The QCML analysis of the Wisconsin Breast Cancer dataset presented in this section 
illustrates how a quantum-geometric representation can be constructed from real-world datasets, 
including complex biomedical data. 
By varying the fluctuation parameter~$w$, the model interpolates between finely resolved and 
coarse-grained geometric descriptions, revealing an effective reduction of the intrinsic dimension 
from $d \approx 7$ at small~$w$ to $d \approx 1$ at large~$w$. 
We have shown how quantum-geometric tools, such as the matrix Laplacian spectrum and eigenmap embeddings, 
can be used to capture
both the global structure and local correlations within the dataset. 
In this work, we used only unsupervised QCML training; the diagnostic labels in the WBC dataset were not used during QCML training, but only for post-training visualization and quantitative evaluation. Since the leading eigenmaps are ordered by smoothness rather than by class variance, incorporating the labels into the loss, for instance, by classifying observables of the kind discussed in Section \ref{sec:eigenmaps}, would be required for the representation to be optimized for discrimination.
}

\section{Discussion}
\label{sec:discussion}

In this work we studied how QCML encodes the data in the form of quantum geometry. We considered a few examples of datasets and showed that the quantum geometry obtained via QCML allows one to uncover some properties of geometric objects underlying the data. In particular, in synthetic examples presented in Sections~\ref{sec:ex-2-spheres}-\ref{sec:ex-conformal-maps} we obtained information on connectivity, intrinsic dimension, quantum geometric tensor, topological invariants, and matrix Laplacian spectra. For the real-life WBC dataset 
\cite{misc_breast_cancer_wisconsin_(diagnostic)_17}
in Section~\ref{sec:ex-WBC},  we used quantum geometric tools to make statements about the connectivity and intrinsic { dimension of the data geometry and to characterize the relative contribution of individual features through their overlaps with the Laplacian eigenmaps.}

Any new way of looking at the data has the potential to reveal hidden patterns and previously unnoticed properties. QCML is inherently geometric in nature. It suggests a quantum geometric perspective for thinking about the data.

We summarize some advantages of applying QCML/quantum geometry for data analysis.
\begin{itemize}
  \setlength{\itemsep}{2pt}

  \item QCML constructs a quantum geometry by learning a configuration of Hermitian matrices and quasi-coherent states from the data, enabling a smooth, compact representation of high-dimensional geometric structures.

  \item Matrix geometries encode complex, nonlinear structures with remarkable efficiency—avoiding lattice artifacts and avoiding the curse of dimensionality at the level of the operator representation when the relevant structure can be captured at small \(N\).

  \item The intrinsic uncertainty of quantum states  can act as a natural form of regularization,  helping to control overfitting and sensitivity to noise.

  \item Unlike classical approaches based on graphs or neural networks, quantum geometry encodes intrinsic notions of distance, curvature, and connectivity via noncommuting observables and quantum states.

  \item Integer-valued topological invariants, such as Chern numbers, can be computed from the learned matrix configuration, providing information on the global structure of the data manifold.

  \item The matrix Laplacian offers a natural spectral tool to extract dominant features, identify disconnected components, estimate the intrinsic dimension, and construct reduced representations of the data.

  \item QCML is naturally suited for implementation on quantum computers. When such platforms are available, they offer the potential for significant computational speedups and scalability advantages.
\end{itemize}

In this work, we focused on unsupervised QCML applied (in synthetic examples) to datasets with underlying smooth geometric structures. Importantly, QCML is not limited to these datasets. The training can be applied to any dataset and will produce some matrix/quantum geometry that may be far from any semiclassical limit. In fact, applications to real datasets often reveal some features that are not smooth. In the context of quantum geometry these features would correspond to ``ultra-quantum'' operators that defy semiclassical treatment (such as spin 1/2 operators or low-rank projectors). These special operators will be necessary for supervised training where they can be used for inference, e.g., in classification problems. 
 
Interestingly, this opens new directions in quantum geometry, particularly in exploring settings where smooth structures coexist with intrinsically quantum features. Another promising object of study is the \emph{geometry of quasi-coherent states} (see Figure~\ref{fig:scheme-QCML}). For example, a quantum geometry can be defined based on the projectors $\ket{x^t}\bra{x^t}$ and infer the properties of the data from this geometry. We believe that more detailed studies in this direction are needed.

From the machine learning perspective, there are several directions that were out of the scope of this work. 
One of them is using quantum distance function (fidelity distance between learned quasi-coherent states) uncertainty structure associated with QCML for classification and clustering algorithms.\cite{rosaler2025supervised} Another, is the use of reduced quantum geometry obtained from matrix Laplacian eigenmaps to compress the data representations. This technique remains largely unexplored and would be a quantum counterpart of known Laplacian eigenmap technique in graph settings \cite{belkin2003laplacian}. Most generally, applications to real-world data—particularly in domains where topology, symmetry, or uncertainty play a central role—could benefit significantly from the spectral tools offered by quantum geometry.

{ An interesting direction for future work, not pursued here, is to apply QCML to
mathematical models of quantum cognition, in the sense reviewed by
\citeonline{BusemeyerBruza2025Book}.}

\section*{Methods}
 \label{sec:methods}

All numerical results in this work were obtained by training matrix configurations $\{X_a\}$ as defined in Eqs.~\eqref{eq:Xkopt} using the loss function in Eq.~\eqref{eq:qcmlloss}. The training process involves iterative updates to both the quasi-coherent states \(|x^t\rangle\) and the matrix configuration \(X\), optimized to minimize the loss over the dataset.
A typical training loop consists, for each epoch, of the following steps:
\vspace{-0.2cm}
\begin{enumerate}
\setlength{\itemsep}{2pt}
    \item Compute the quasi-coherent states \(|x^t\rangle\) for all data points \(x \in \mathcal{X}\) (or a mini-batch).
    \item Evaluate the loss function in Eq.~\eqref{eq:qcmlloss} and compute its gradients with respect to the matrix configuration \(X\).
    \item Update \(X\) using gradient descent via the Adam optimizer.
\end{enumerate}
All training routines were implemented in PyTorch~\cite{paszke2019pytorch}. Spectral quantities such as matrix Laplacian spectra and eigenmaps were computed from the trained matrix configurations using standard linear algebra routines from NumPy and SciPy. Appendix~\ref{app:examples-summary} provides a summary of the parameters used for numerical computations.

{ The dominant computational cost of QCML training is the repeated construction
and diagonalization of the \(N\times N\) displacement Hamiltonian. For each
data point, constructing the Hamiltonian requires \(O(DN^2)\) operations,
while dense diagonalization requires \(O(N^3)\). The approximate cost per
epoch is therefore
\[
    O\!\left(T(DN^2+N^3)\right),
\]
up to minibatching, automatic differentiation, and hardware-dependent
overhead.

As an empirical timing check, we repeated the conformal-map training for
\(T=2000,4000\), \(D=200,400\), and \(N=8,16\), using \(100{,}000\) epochs
and minibatches of size \(100\). The original configuration
\((T,D,N)=(2000,200,8)\) required \(5.44\) minutes. Across the tested
configurations, the training times ranged from \(5.44\) to \(24.73\) minutes.
These measurements were obtained using CPU execution on an Apple M4 processor
with \(24\) GB of unified memory. They are single-run wall-clock measurements
and exclude dataset generation and final full-dataset diagnostics.}

{
\section*{Code and data availability}

The synthetic datasets, final trained Hermitian operators, selected derived
quantities, and post-training analysis and figure-generation scripts used in
this work are available in the public repository\cite{hogan2026quantumgeometrydata}. The repository contains the materials
needed to reproduce the figures for the uniform-sphere validation and the
examples in Section~\ref{sec:examples}, including the two-spheres stability study. The Wisconsin
Breast Cancer dataset is publicly available from the source cited in the text.\cite{misc_breast_cancer_wisconsin_(diagnostic)_17}
For WBC, the repository includes the six operator sets and saved response spectra used for Table~\ref{tab:intrinsic-dimension}, together with the analysis scripts and operator-dependent classifier outputs.
The QCML training implementation is available from the authors upon reasonable
request.
}

\section*{Acknowledgments}

We thank Haim Bar for his valuable insights regarding the Wisconsin Breast Cancer dataset.

We gratefully acknowledge Qognitive for bringing the authors together and providing critical support. In the context of limited public funding, we especially value the opportunity for collaboration between researchers from the creative technology sector and academic institutions. This partnership enabled foundational research at the intersection of statistical theory, machine learning, and mathematical physics. This research, which forms the basis for methodological advances in artificial intelligence, would not have been possible without Qognitive’s support.



\newpage

\begin{appendix}

\section{Basic examples of quantum geometries}
\label{app:QG-basic-examples}

We consider a few basic examples of quantum geometries discussed in the literature \cite{steinacker2021quantum}.

\paragraph{Commuting matrix configurations.}

For a matrix configuration with vanishing commutators, \( [X_a, X_b] = 0 \), there exists a common eigenbasis \( |i\rangle \) such that
\[
X_a = \sum_i x_a^{(i)} |i\rangle\langle i|.
\]
This defines \( N \) points \( x_a^{(i)} \in \mathbb{R}^D \) in feature space. Since the displacement Hamiltonian is diagonal in this basis, the quasi-coherent state \( |x\rangle \) is simply the eigenstate \( |i\rangle \) corresponding to the point closest to \( x \). The ground state energy \( \lambda(x) \) is then half the squared distance from \( x \) to the nearest \( x^{(i)} \).

Thus, for a given dataset, the commuting matrix configuration that minimizes the total ground state energy effectively recovers the K-means clustering into \( N \) centroids \( x_a^{(i)} \in \mathbb{R}^D \).

While simple and intuitive, commuting configurations suffer from lattice artifacts and lack smoothness. In contrast, quantum geometries based on noncommuting \( N \times N \) matrices yield fuzzy but smooth approximations of continuous spaces.

\paragraph{The fuzzy sphere and higher-dimensional generalizations.}

The fuzzy sphere \( S^2_N \) (Ref.~\citeonline{madore1992fuzzy}) is a prototypical example of a nontrivial quantum geometry. It is defined by a matrix configuration consisting of three generators of the \( N \)-dimensional irreducible representation of \( SU(2) \), satisfying
\begin{align}
    \sum_{a=1}^3 X_a X_a = \tfrac{1}{4}(N^2 - 1)\,\one, \qquad
    [X_a, X_b] = i \varepsilon_{abc} X_c.
\end{align}
The normalization, i.e., the radius of the sphere, is conventional and can be adjusted via \( X_a \to \alpha X_a \).
The spectrum of the matrix Laplacian is found to be
$\mathrm{spec}(\Delta) = \{l(l+1), \; l = 0, \ldots, N-1\}$, with multiplicities $2l+1$.

As shown in Ref.~\citeonline{steinacker2021quantum}, Sec.~6.1, the associated quasi-coherent states \( |x\rangle \) are given by \( SU(2) \) rotations of the highest weight state. The minimal ground state energy is \( \lambda(x) = \tfrac{1}{4}(N - 1) \), attained for points \( x \) on a sphere of radius \( |x| = \tfrac{1}{2}(N - 1) \). The full set of quasi-coherent states recovers the sphere:
\begin{align}
    \mathcal{M} = \{ |x\rangle \;|\; x \in \mathbb{R}^3 \setminus \{0\} \} \cong S^2.
\end{align}
In the minimal case \( N = 2 \), this reduces to the Bloch sphere. Here, the states \( |x\rangle \) coincide for all \( x \) along a half-line from the origin, so the map \eqref{targetspace-map} becomes a projection from \( \mathbb{R}^3 \setminus \{0\} \) onto the nearest point on \( S^2 \). The origin is the unique point where the displacement Hamiltonian is (maximally) degenerate.

Using these quasi-coherent states and the associated symbol maps, one obtains a natural correspondence
\begin{align}
    \mathrm{Mat}(N) \longleftrightarrow L^2(S^2),
\end{align}
which is compatible with \( SO(3) \) symmetry and truncated at angular momentum \( \ell_{\text{max}} = N - 1 \).

\paragraph{Minimal fuzzy \( \mathbb{C}P^{N-1} \).}

The fuzzy sphere is the simplest case in a broader class of higher-dimensional, maximally symmetric quantum spaces, which can be interpreted as quantized coadjoint orbits of semi-simple Lie groups \( G \). These are constructed from matrix configurations \( \{X_a, \ a = 1, \ldots, D\} \), where the \( X_a \) are generators of a unitary irreducible representation of \( G \). As in the previous example, the associated quasi-coherent states allow one to reconstruct the underlying classical manifold \( \mathcal{M} \) from the quantum data \cite{steinacker2024quantum}. 

A natural higher-dimensional analog of the Bloch sphere is given by the minimal fuzzy complex projective space \( \mathbb{C}P^{N-1} \), constructed from the matrix configuration
\begin{align}
\label{fuzzy-CPN-min}
    X_a = \lambda_a, \qquad a = 1, \ldots, N^2 - 1,
\end{align}
where \( \lambda_a \) are the (generalized) Gell-Mann matrices—i.e., the generators of the fundamental (or anti-fundamental) 
representation of \( SU(N) \). This yields a matrix geometry in \( \mathbb{R}^{N^2 - 1} \) with underlying dimension \( 2(N-1) \), matching that of the classical complex projective space \( \mathbb{C}P^{N-1} \), which is a homogeneous space of \( SU(N) \).

As in the fuzzy sphere case, the set of quasi-coherent states \( |x\rangle \) recovers the classical manifold:
\begin{align}
   \mathcal{M} = \{ |x\rangle \;|\; x \in \mathbb{R}^{N^2 - 1} \setminus \{0\} \} \cong \mathbb{C}P^{N-1}.
\end{align}
The map in \eqref{targetspace-map} again defines a projection from feature space onto the closest point on this manifold. This leads to a natural correspondence
\begin{align}
    \mathrm{Mat}(N) \longleftrightarrow L^2(\mathbb{C}P^{N-1}),
\end{align}
compatible with the action of \( SU(N) \). In the minimal case (i.e., for the fundamental representation), this correspondence captures polynomial functions of degree one on \( \mathbb{C}P^{N-1} \); higher-degree modes appear when higher representations are used \cite{balachandran2002fuzzy}.
For (general) fuzzy $\C P^{N-1}$, the spectrum of the
matrix Laplacian is found to be
$\mathrm{spec}(\Delta) = \{2k(k+N-1), \; k = 0, \ldots, k_{\rm max}\}$, with multiplicities $\sim k^{d-1}$ for large $k$, where $d=2(N-1)$ is the dimension of $\C P^{N-1}$. This is consistent with Weyl's law.

\paragraph{The fuzzy torus.}

A further basic example of a quantum geometry is the fuzzy torus $T^2_N$. It is defined by a matrix configuration consisting of four matrices $X_a$ defined by
\begin{align}
X_1 + i X_2 := U, \qquad X_3 + i X_4 := V \ 
\label{torus-embedding}
\end{align}
where $U,V$ are unitary "clock" and "shift" operators satisfying 
\begin{align}
U V = q V U,  \qquad U^N = V^N = 1 \
\label{UV-CR-torus}
\end{align}
with $q=e^{2\pi i/N}$. The $X_a$ can be interpreted as quantized flat embedding map \eqref{embedding-map2} of the 2-dimensional (Clifford) torus $T^2$ into $\R^4$. 
The $U(1) \times U(1)$ symmetry of the classical $T^2$ is broken to $\Z_N \times \Z_N$ in the fuzzy torus, and a quantum geometry point cloud would show $N$ slight bumps, corresponding to the quantum cells. The abstract quantum space $\cM$ turns out to be ``oxidized'' i.e. somewhat thickened, as in generic quantum geometries.
Nevertheless, the expected geometrical features are reproduced to a good approximation for sufficiently large $N$.
The spectrum of the matrix Laplacian is  $\mathrm{spec}(\Delta) = c_N \{[n]_q^2 + [m]_q^2, \; n,m = -\frac N 2,\ldots,\frac N2\}$, assuming that $N$ is even. Here 
\begin{align}
[n]_q :=  \frac{\sin(n\pi/N)}{\sin(\pi/N)} \ \sim \  n \ ,
\end{align}
so that the classical spectrum is recovered sufficiently far below the cutoff $N/2$.

\section{Spectral and variational structure of the matrix Laplacian}
\label{app:matrix-laplacian}

Given a matrix configuration \( \{X_a\} \subset \Mat(N) \), we define the matrix Laplacian as a second-order operator on \( \Mat(N) \),
\begin{align}
\label{eq:matrixLaplacian2}
    \Delta = \sum_a [X_a, [X_a, \cdot]].
\end{align}
This operator is Hermitian and positive semi-definite, and its spectrum encodes both algebraic and geometric information about the matrix configuration.

In the semi-classical regime, \( \Delta \) approximates the Laplace--Beltrami operator on an emergent manifold \( \mathcal{M} \subset \mathbb{C}P^{N-1} \),
\begin{align}
\label{laplacian-geometric}
    \Delta \sim e^{2\varphi} \Delta_G,
\end{align}
where \( \varphi \) is a dilaton field and \( G \) is an effective metric that may differ from the quantum metric \( g \) associated with the quasi-coherent state construction \cite{steinacker2024quantum, steinacker2010emergent}.

A natural quadratic form associated with the Laplacian is the \emph{energy functional},
\begin{align}
    E[Y] = \langle Y, \Delta Y \rangle = \mathrm{Tr}(Y \Delta Y) = \sum_a \mathrm{Tr}([X_a, Y]^\dagger [X_a, Y]),
\end{align}
which measures the extent to which \( Y \in \Mat(N) \) fails to commute with the observables \( X_a \). Diagonalizing \( Y = \sum y_i |y_i\rangle\langle y_i| \), one finds:
\begin{align}
    \nabla^a Y = [X_a, Y] = \sum_{i \neq j} (y_i - y_j) x^a_{ij} |y_i\rangle\langle y_j|, \qquad
    E[Y] = \sum_{a, i \neq j} (y_i - y_j)^2 |x^a_{ij}|^2,
\end{align}
where $x_{ij}^a=\bra{y_i}X_a\ket{y_j}$.
Minimizing \( E[Y] \) under the constraint \( \|Y\|_2 = 1 \) yields the eigenmaps of the matrix Laplacian:
\begin{align}
    \Delta Y_k = \lambda_k Y_k, \qquad
    \lambda_k = \min \left\{ E[Y] \;\middle|\; \|Y\|_2 = 1,\ \langle Y, Y_j \rangle = 0\ \text{for } j < k \right\}.
\end{align}

The lowest eigenvalues and their corresponding eigenmaps provide optimally flat representations of the quantum geometry and can be used to construct low-dimensional embeddings. In particular, they generalize the concept of Laplacian eigenmaps \cite{belkin2003laplacian} to noncommutative spaces.

For commuting matrix configurations, the lowest \( N \) eigenvalues vanish, with eigenmaps corresponding to diagonal matrices. In this case, the first nontrivial eigenmaps reflect adjacency relations between discrete points, linking this setting to classical graph Laplacians. In contrast, for irreducible configurations, the non-zero spectrum reflects intrinsic curvature and dimension.

The full eigenvalue distribution provides several diagnostics:
\begin{itemize}
\item Zero modes signal disconnected components in reducible configurations (see Section~\ref{sec:eigenmaps}). The zero modes of the matrix Laplacian are used to identify disconnected components in the quantum geometry of the data. In reducible matrix configurations, where the observables have a block-diagonal structure, the Laplacian has multiple zero eigenvalues in addition to the trivial one for the identity matrix. Each zero mode acts as a projector onto an irreducible block, effectively separating the dataset into distinct components. By analyzing the eigenspace associated with these zero (or near-zero) eigenvalues, it is possible to algorithmically identify and extract the disconnected parts of the quantum manifold.

\item Weyl's law describes the asymptotic distribution of eigenvalues of the (matrix) Laplace operator\cite{steinacker2010emergent}. Specifically, it relates the number of eigenvalues less than or equal to a threshold $\lambda$ to the geometric dimension of the underlying space. This method is particularly useful in the context of numerical studies of fuzzy geometries, spectral manifolds, or manifold learning, where the spectrum of a Laplacian (or Laplace-like operator) is available and one wishes infer the effective dimension of the space being approximated, even if the space itself is not presented in classical geometric form. Let $N(\lambda)$ denote the counting function, the number of Laplacian eigenvalues less than or equal to $\lambda$. Then Weyl's law states that $
N(\lambda) \sim C \cdot \lambda^{d/2}, $ as $\lambda \to \infty,$
where $d$ is the intrinsic dimension of the manifold, and $C$ is a constant that depends on geometric quantities such as volume.

This asymptotic relation provides a practical tool for estimating the intrinsic dimension $d$ of a manifold from spectral data. By taking the logarithm of both sides of the asymptotic relation, we obtain $ \log N(\lambda) \sim \frac{d}{2} \log \lambda + \log C.$  This equation has the form of a straight line in a log--log plot of $N(\lambda)$ versus $\lambda$, with slope equal to $d/2$. Therefore, if one plots $\log N(\lambda)$ against $\log \lambda$, the slope of the resulting line provides an estimate of the intrinsic dimension.
\item The spectrum encodes geometric information analogous to the classical Laplace–Beltrami operator \cite{connes2010noncommutative}. Low eigenvalues reflect global structures, such as connected components, while the growth rate of higher eigenvalues reveals the intrinsic dimension of the space. In the context of noncommutative geometry, these spectral data replace traditional notions of distance and curvature, allowing geometry to emerge solely from the encoded subspace. Therefore, the spectrum becomes a fundamental tool for examining shape, connectivity, and dimensionality in quantum spaces. 
\end{itemize}

In summary, the matrix Laplacian serves as a central tool for probing curvature, reducibility, and dimensionality of quantum spaces, with spectral properties supporting variational principles that parallel classical differential geometry. The Section~\ref{sec:eigenmaps} discusses how to exploit these properties for model reduction and dimensionality estimation.

\section{Structural properties of matrices}
 \label{sec:structural}

It is useful to distinguish different types of irreducible quantum configurations:
\begin{itemize}

\item Classical, ie, commutative matrix configurations $[X_a,X_b] = 0 \ \forall a,b$.
 This corresponds to lattices, with vertices corresponding to the common eigenvectors $|i\rangle$.
 The eigenmaps of $\Delta$ 
are given by $Y_{ij} = |i\rangle\langle j| + h.c.$ and satisfy $\Delta Y_{ij}  = \sum_a(x^a_i - x^a_j)^2 Y_{ij}$.
In particular, there are $N$ zero modes given by $|i\rangle\langle i|$.

\item Almost-commutative matrix configurations corresponding to semi-classical quantum geometries. 
The crucial property is that the commutator
is much smaller than the matrices:
$\|[X_a,X_b]\|_2 \ll \|X_a X_b\|_2$,
so that they can be "almost" simultaneously diagonalized by the quasi-coherent states.
Such matrix configurations can typically
be interpreted in terms of quantum geometry.
This is the most interesting case, 
leading to an almost-continuous spectrum of $\Delta$ following some power law.

 \item Deep quantum configurations,
characterized by 
$\|[X_a,X_b]\|_2 = O(\|X_a X_b\|_2)$ and/or $E(X) = O(\lambda_{max}) \|X\|_2^2$.
This holds e.g. for random matrices, which
satisfy $\|X_a X_b\|_2 \sim \frac 12\|[X_a,X_b]\|_2$; moreover, their
spectral distribution near the cutoff is very distinct from geometric configurations (power law versus $\sqrt{x}$).
These are not interesting from the point of view of quantum geometry.
    
\end{itemize}

We can also distinguish different types of individual observables $Y$ (normalized as $\|Y\|_2 =1$):

\begin{itemize}

\item 
{\bf Local = semi-classical} observables in the IR range of the Laplacian, with energy $E(Y) \approx \lambda_0$.
These can be interpreted as slowly varying local functions on the data manifold $\cM$.
They  can be written in the form $\phi = \int \varphi(x) |x\rangle\langle x|$, according to the semi-classical relation \eqref{class-Quant-rel}.

\item 
{\bf Non-local = deep quantum} observables such as $Y = |x\rangle\langle y| + h.c.$. For these we expect 
$\langle x_i| Y |x_i\rangle \approx 0$ for all data points $x_i$, 
and the semi-classical relation \eqref{class-Quant-rel} fails completely.
They are in the UV regime of the Laplacian and high energy
 $E(Y) \approx  \lambda_{max}$.
 These operators may be interesting to measure non-local correlations in data space via $\Tr(\rho Y \rho Y) \neq 0$
 where  $\rho = \frac 1N\sum |x_i\rangle\langle x_i|$ is the data density matrix,
 or to characterize non-classical properties.

\item 
Operators \( Y^\dagger Y = Y^2 \) are positive definite and normalized. They can be viewed as density matrices and characterized by their (von Neumann) entropy,
$S[Y^2] = -\Tr(Y^2 \ln Y^2)$.
Operators with low entropy are not semi-classical but may still be nearly local, such as projectors or classifying operators.

\end{itemize}

\section{Summary of examples used}
 \label{app:examples-summary}

\begin{table}[htbp]
\centering
\setlength{\tabcolsep}{2.5pt}
\caption{Summary of examples used}
\label{tab:examples}
\begin{tabular}{|l|c|c|c|p{3.5cm}|c|c|p{2.1cm}|c|}
\hline
\textbf{Example} & \textbf{Section} & \textbf{Figure(s)} & \textbf{\# Points} & \textbf{Distribution} & \textbf{\( D \)} & \textbf{\( N \)} & \textbf{\( w \)} & \textbf{Added Noise} \\
\hline
Fuzzy sphere ($S^2_N$) & \ref{sec:QCML}, \ref{sec:matrix-Laplacian} & \ref{fig:1-sphere-N4-E20000}, \ref{fig:laplacian-spectrum-fuzzy-sphere} & 1000 & Uniform on sphere & 3 & 4 & 0.1 & 0 \\
\hline
Two noisy spheres & \ref{sec:ex-2-spheres} & \ref{fig:2spheres-QCML}, \ref{fig:2spheres-Cloud-uncertainty} & 2000 & Uniform on 2 spheres & 3 & 8 & 0.1 & 0.1 \\
\hline
Non-uniform sphere & \ref{sec:ex-nonuniform-sphere} & \ref{fig:1spheres-nonuniform-QCML}, \ref{fig:1spheres-nonuniform-Cloud-uncertainty} & 2000 & \leavevmode\( \propto (1+\cos\theta)^3 \)\newline per unit surface area & 3 & 8 & 0.1 & 0.1 \\
\hline
Conformal maps & \ref{sec:ex-conformal-maps} & \ref{fig:conformal-map-examples}, \ref{fig:conformal-metric-spectral}, \ref{fig:XY-overlap}, \ref{fig:conformal-eigenmap-colored}, \ref{fig:conformal-3D-colored} & 2000 & Random M\"obius transforms of 100 2D points & 200 & 8 & 0.1 & 0 \\
\hline
Breast cancer (WBC) & \ref{sec:ex-WBC} & \ref{fig:WBC}, \ref{fig:WBC2}, \ref{fig:WBC3} & 569 & Empirical (diagnostic features) & 30 & 8 & \leavevmode0.02, 0.05, 0.1\newline 0.2, 0.5, 0.8 & 0 \\
\hline

\end{tabular}
\end{table}
{
For reproducibility, all synthetic datasets and training runs used for the
figures were generated with random seed $42$, unless otherwise stated. The
learned Hermitian operators were initialized by sampling their independent
real parameters from a zero-mean, unit-variance normal distribution and scaling
the resulting matrices by $1/\sqrt{2N}$. Optimization used Adam with
$\beta_1=\beta_2=0.99$, AMSGrad, and zero weight decay. Training used fixed
epoch budgets rather than convergence-based early stopping. The random seeds,
learning rates, minibatch sizes, numbers of epochs, Hilbert-space dimensions
$N$, and values of $w$ used for the reported examples are provided with the
corresponding scripts and trained operators in the accompanying repository\cite{hogan2026quantumgeometrydata}.
}

\section{Glossary}
 \label{app:glossary}

\begin{description}[leftmargin=2.5cm, style=nextline]

\item[Feature operator; observable \( X_a \)] 
A learned Hermitian matrix of size \( N \times N \) representing the \( a \)-th data feature in QCML. The collection \( \{X_a\} \) encodes the quantum geometry of the dataset.

\item[Quasi-Coherent State \( \ket{x^t} \)]
A normalized quantum state corresponding to the \( t \)-th data point. Learned jointly with the observables \( X_a \), these states provide a soft encoding of classical input features.

\item[QCML (Quantum Cognition Machine Learning)]
A framework that represents data using quantum structures. Data features are mapped to observables \( \{X_a\} \), and data points to quasi-coherent quantum states \( \{\ket{x^t}\} \). Both are learned through an optimization process that minimizes a loss function combining displacement and uncertainty. QCML enables geometric and topological analysis of data via quantum geometry tools.

\item[QCML Loss Function]
The total loss optimized during QCML training $L = d^2 + w \sigma^2 $,
where \( d^2 \) measures average displacement and \( \sigma^2 \) represents the quantum uncertainty. The weight \( w \) balances the two contributions.

\item[QCML Cloud]
A geometric point cloud in \( \mathbb{R}^D \) defined by the expectation values $\tilde{x}_a^t := \bra{x^t} X_a \ket{x^t}$ across data points and features. The QCML cloud captures a data-driven embedding that combines learned observables and states.

\item[Quantum Geometry Cloud]
A visualization of the quantum geometry based on expectation values of the observables \( \{X_a\} \) under quasi-coherent states of a random point cloud. Unlike the QCML cloud, it depends only on the learned operators and not on the quasi-coherent states. Used, e.g., in Figure~\ref{fig:2spheres-Cloud-uncertainty}.

\item[Quantum Geometry]
The geometric structure defined by the set of learned observables \( \{X_a\} \). It includes features such as metric structure, curvature, and topological invariants derived from operator relationships.

\item[Geometry of Quasi-Coherent States]
The geometric structure inferred from the projectors \( \ket{x^t}\bra{x^t} \). While not part of the quantum geometry proper, it provides complementary insight into the data structure.

\item[Matrix Laplacian \( \Delta \)]
An operator acting on matrices, defined as $\Delta = \sum_a [X_a, [X_a, \cdot]]$. It generalizes the classical Laplace-Beltrami operator and is used to probe spectral and topological properties of quantum geometry.

\item[Matrix Laplacian Eigenmaps]
The eigenmatrices $Y_i$ of the matrix Laplacian found from the eigenvalue problem $\Delta Y_i =\mu_i Y_i$. The leading eigenmaps (with { smaller} $\mu_i$) reveal large-scale geometric features of the learned data manifold.

\item[Displacement Hamiltonian]
An operator of the form $H(x) = \tfrac{1}{2}\sum_a ( X_a -  x_aI_N)^2$ used to define the quasi-coherent state \( \ket{x} \) associated with the point $x\in \mathbb{R}^D$. 

\item[Quantum Metric (Quantum Geometric Tensor)]
A local geometric tensor derived from the variation of quantum states. Its spectrum encodes the intrinsic dimensionality and curvature properties of the quantum geometry.

\item[Intrinsic Dimension]
The effective dimensionality of the data manifold. In QCML, it can be estimated from the spectrum of the response matrix or the matrix Laplacian.

\end{description}

\end{appendix}

\end{document}